%% file: main.tex
\documentclass[letterpaper, 10 pt, journal, twoside]{ieeetran}

\DeclareMathAlphabet{\mathcal}{OMS}{cmsy}{m}{n}

\usepackage{lmodern}
\usepackage{amsmath} 
\usepackage{amsfonts}  

\usepackage{graphics} 
\usepackage{graphicx} 
\usepackage{epsfig} 
\usepackage{bm} 
\usepackage{times} 
\DeclareMathOperator*{\argmin}{arg\,min}  
\usepackage{amssymb}  
\usepackage{bm}
\usepackage{algorithm}
\usepackage{algpseudocode}
\usepackage{mathtools}
\usepackage{listings}
\usepackage{subcaption}
\usepackage{multirow}
\usepackage{hyperref}
\usepackage[table,xcdraw]{xcolor}
\usepackage{optidef}

\newcommand{\R}{\mathbb{R}}

\newcommand{\br}[1]{\mathbf{\bm{#1}}}
\newcommand{\tp}{\top}
\newcommand{\bmat}[1]{\begin{bmatrix} #1 \end{bmatrix}}

\begin{document}

\title{\LARGE \bf
Online Learning-Based Inertial Parameter Identification of Unknown Object for Model-Based Control of Wheeled Humanoids
}

\author{Donghoon Baek$^{1}$, Bo Peng$^{2}$, Saurabh Gupta$^{3}$, and Joao Ramos$^{1,2}$
\thanks{This work is supported by University of Illinois, 1-200250-917015-917396.}
\thanks{The authors are with the $^1$ Department of Mechanical Science and Engineering and the $^2$ Department of Electrical and Computer Engineering and the $^3$ Department of Computer Science at the University of Illinois at Urbana-Champaign, USA.{\tt\small dbaek4@illinois.edu}} 
}

\maketitle

\begin{abstract}
Identifying the dynamic properties of manipulated objects is essential for safe and accurate robot control. Most methods rely on low-noise force-torque sensors, long exciting signals, and solving nonlinear optimization problems, making the estimation process slow. In this work, we propose a fast, online learning-based inertial parameter estimation framework that enhances model-based control. We aim to quickly and accurately estimate the parameters of an unknown object using only the robot's proprioception through end-to-end learning, which is applicable for real-time system. To effectively capture features in robot proprioception affected by object dynamics and address the challenge of obtaining ground truth inertial parameters in the real world, we developed a high-fidelity simulation that uses more accurate robot dynamics through real-to-sim adaptation. Since our adaptation focuses solely on the robot, task-relevant data (e.g., holding an object) is not required from the real world, simplifying the data collection process. Moreover, we address both parametric and non-parametric modeling errors independently using \textit{Robot System Identification} and \textit{Gaussian Processes}. We validate our estimator to assess how quickly and accurately it can estimate physically feasible parameters of an manipulated object given a specific trajectory obtained from a wheeled humanoid robot. Our estimator achieves faster estimation speeds (around 0.1 seconds) while maintaining accuracy comparable to other methods. Additionally, our estimator further highlight its benefits in improving the performance of model based control by compensating object's dynamics and re initializing new equilibrium point of wheeled humanoid.


\end{abstract}

\begin{IEEEkeywords}
Inertial Parameter Estimation, Real-to-Sim Adaptation, Representation Learning
\end{IEEEkeywords}

\input{1_introduction}

\input{2_related_work}

\input{3_background}

\input{4_method}

\input{5_experiment}
\input{6_results_and_discussion}
\input{7_conclusion}



\bibliographystyle{IEEEtran}
\bibliography{main.bib}


\end{document}

%% file: 1_introduction.tex
\section{INTRODUCTION}
\label{S:1}
Collaborative robots (e.g., Humanoid and manipulator) have become increasingly prevalent with great potential in various fields such as manufacturing, healthcare, and even disaster response \cite{purushottam2022hands,nadeau2022fast}. To facilitate seamless and safe human-robot collaboration, these robots need an accurate understanding of the physical properties of the objects they interact with. Understanding an object's inertial parameters (mass, center of mass, moment of inertia) enables robots to interact more robustly and adaptively. For instance, insufficient information may lead to excessive or insufficient force application, causing slipping or damage during manipulation tasks.


Identifying the inertial parameters of an unknown object with a humanoid robot is challenging due to several inherent factors: (1) noisy signals from force-torque sensors, (2) the need for long excitation trajectories to collect sufficient data, and (3) solving nonlinear constraint optimization for physically feasible parameters. These factors inherently slow down the estimation process, and the use of force-torque sensors and cameras is not always accessible.



\begin{figure}[t!]
\begin{center}
\includegraphics[width=0.65\linewidth]{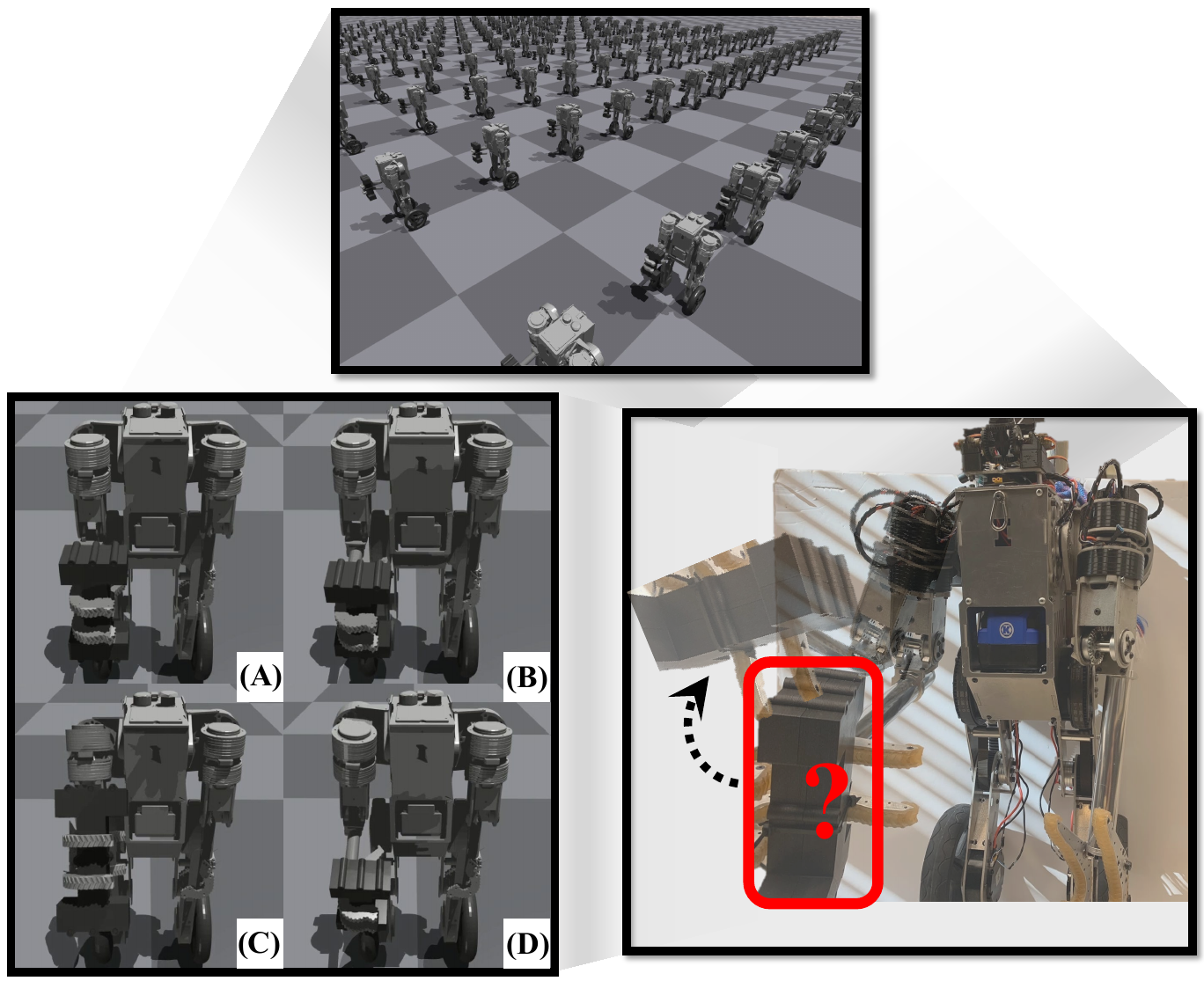}
\end{center}
\caption{\textbf{Conceptual overview of the proposed method.} Thousands of wheeled humanoid robots, SATYRR, in a simulator shake an object to identify its inertial parameters. Different object dynamics impact the robot’s proprioception in varied ways. The effect of different object dynamics on the robot's proprioception closely mirrors real-world conditions due to effective real-to-sim adaptation. During training, the object's inertial parameters can be sampled either randomly or within specific shape boundaries.}
\vspace{-1.5em}
\label{fig1}
\end{figure}

In this paper, we propose a learning-based approach to quickly identify the inertial parameters of an unknown object, enhancing the performance of a model-based controller. Our method uses a data-driven regression model relying solely on proprioception errors influenced by object dynamics, not requiring long excitation signals and constrained nonlinear optimization. While using proprioception to identify parameters is not a new idea, \textit{estimating the physically feasible full inertial parameters of unknown object in an end-to-end manner is a novel approach}. The main challenge is efficiently capturing the object's dynamic effects in the robot's proprioception while successfully transferring this information from simulation to the real world. For example, proprioception is influenced by the robot's own dynamics, the object's dynamics, and even the \textit{reality gap}. To achieve this, we identify an accurate model of the robot to minimize its reality gap via real-to-sim adaptation, leaving only the effect of object dynamics on the robot's proprioceptive error. Unlike other approaches, we focus only on robot dynamics, eliminating the need to collect task-relevant data from the real world (e.g., the robot does not need to hold an object in the real world). Our adaptation addresses both parametric and non-parametric modeling errors using a combination of \textit{Robot System Identification} (SysID) and \textit{Gaussian Processes} (GPs). The data collection and training of the estimator are conducted offline and then transferred to the real world with zero-shot adaptation, ensuring fast inference times without additional iterations or tuning. We evaluate the estimation speed and accuracy of our estimator by benchmarking it against previous methods and also assessing the effects of the components in our adaptation method. Results show that our estimator can identify an object's inertial parameters within 0.1 seconds while achieving accuracy comparable to other baselines. Additionally, we show benefits of our method in improving the performance of model-based control by complementing object dynamics in manipulation tracking and delivering object locomotion tasks, achieving 36\% and 65\% performance improvements in each task, respectively. 

Our summarized contributions are: (1) a fast inertial parameter estimation framework using an end-to-end data-driven model; (2) effective handling of both parametric and non-parametric modeling errors to identify accurate robot dynamics, reducing its \textit{reality gap}; (3) demonstration on physical hardware, highlighting the benefits of accurate inertial parameters in enhancing model-based controller performance.

%% file: 2_related_work.tex
\section{RELATED WORK}
\label{Related:1}
\subsection{System Identification}
Research on inertial parameter identification of rigid bodies has a rich history \cite{atkeson1986estimation}. Since the dynamics model of a multi-body system is linear with respect to the inertial parameters, a linear least squares method has been widely used for this identification, whether through offline or adaptive means. However, this traditional approach has several limitations that must be further considered.

Without including constraints in the optimization process, not all combinations of parameters correspond to the physical system. Some studies have addressed this by incorporating constraints like custom manifold optimization \cite{traversaro2016identification}, linear matrix inequalities (LMIs) \cite{wensing2017linear}, and Riemannian metrics \cite{lee2018geometric}. These methods ensure physical consistency but are time-consuming (over 6 seconds) and require prior object knowledge (e.g., shape, CAD data). Our approach implicitly enforces dynamic consistency using regularization, achieving parameter estimation, including trajectory generation, in under 0.1 seconds.

Force-torque (FT) sensors are essential for determining an object's inertial parameters but are often hindered by weight, cost, and noisy outputs \cite{sundaralingam2021hand}. Alternatives like neural network-based force/torque estimation still require motor torque sensors \cite{yilmaz2020neural}. A recent study \cite{lao2023learning} used encoder discrepancies and attention mechanisms to estimate mass and center of mass (COM) without FT sensors. Our approach builds on this by estimating the full set of inertial parameters without relying on torque/force sensors.

Learning-based inertial parameter estimation has been actively studied, often utilizing extensive visual datasets like images and videos \cite{standley2017image2mass, wu2015galileo}. While accessing a visual dataset is not challenging, interaction with the object is crucial for estimating dynamics due to limited information like density. Our approach only relies on interacting with an object, without using cameras. 

\subsection{Sim-to-Real Transfer}
With the recent advancements in reinforcement learning (RL), sim-to-real transfer has gained significant attention and is becoming increasingly crucial. Domain randomization, optimizing simulation parameters, and reducing parameter distribution \cite{chebotar2019closing, tiboni2023dropo, du2021auto, tsai2021droid, allevato2020tunenet, muratore2022neural, gruner2024pseudo} have been studied to reduce the \textit{reality gap}. While these methods yield promising results, they require multiple iterations, making them unsuitable for real-time applications. Moreover, task-relevant data is often required in the training process, which is challenging to obtain, especially for the inertial parameters of an object. Similar work with ours can be found in \cite{wu2012semi} regarding using Gaussian Processes for robot system identification. In contrast, we employed these methods independently to address both parametric and non-parametric modeling errors. Some literature demonstrates the capability of estimating inertial parameters with deep neural networks \cite{yu2017preparing}. While this method is fast and effective, it requires to use a pre-trained RL policy.

In the control perspective, approaches like domain randomization and privileged learning \cite{li2023robust, hwangbo2019learning} are designed for zero-shot policy transfer to the real world. While effective in reducing the \textit{reality gap} for control, they have limited performance in matching state trajectories, which impacts parameter estimation accuracy \cite{li2023robust}.

%% file: 3_background.tex
\section{BACKGROUND}
\label{B:1}
\subsection{Inertial Parameter Identification}
The inertial parameters of $i$'th rigid-body are collected as $\br{\phi}_i=[m_i, \mathbf{c}_i^\tp, I^{xx}_i,I^{yy}_i,I^{zz}_i,I^{xy}_i,I^{yz}_i,I^{zx}_i]^\tp \in \mathbb{R}^{10}$, where $m \in \mathbb{R}$ is the mass, $\mathbf{c}=[c_x,c_y,c_z]^\tp \in \mathbb{R}^3$ is the the center of mass (COM) position, and $I^{(\cdot)}$ are the moments and the products of inertia. 
In the classical way, the inertial parameters of $n_s$-rigid-body system can be estimated by classical linear regression model $\br{Y}(\br{a}, \br{\omega}, \dot{\br{\omega}})$ using the rigid body dynamics, 

\begin{equation}
 \bmat{\br{f}\\ \br{\tau}} = \bmat{m\br{a}  \\ [\br{\omega}]\br{I}\br{\omega} + \br{I}\dot{\br{\omega}}} = \br{Y} \phi.
\label{eqn_newton_euler_1}
\end{equation}
where the symbolic $\br{f} \in \mathbb{R}^3$ and $\br{\tau} \in \mathbb{R}^3$ denote forces and torques, respectively. The term $m \in \mathbb{R}^+$ denotes the mass, $\br{a} \in \mathbb{R}^3$ is the linear acceleration, $\br{\omega} \in \mathbb{R}^3$ is the angular velocity, $\dot{\br{\omega}} \in \mathbb{R}^3$ is the angular acceleration, and the bracket denotes $[\,\cdot\,]$ the skew-symmetric representation of a vector. 

A simplistic estimation of inertial parameters with least-square regression may result in physically invalid parameters \cite{wensing2017linear}. Hence, the regression could be enforced with physical consistency \cite{traversaro2016identification} as follows,   
\begin{mini!}|s|
{\mathbf{R}, \mathbf{J}, \mathbf{c}, m}{\sum_{n} \left\| \mathbf{Y}^{(n)} \pi(\mathbf{R}, \mathbf{J}, \mathbf{c}, m) - \mathbf{\tau}^{(n)} \right\|^2}
{\label{opti}}{}
\addConstraint{\mathbf{R} \in SO(3)
\label{opti:rotation}}
\addConstraint{m > 0, \ J_i > 0, \ i = 1,2,3
\label{opti:mass}}
\addConstraint{J_1 + J_2 + J_3\ge 2J_k, k=1,2,3 \label{opti:tri}}{}
\end{mini!}
to enforce the manifold constraint $\eqref{opti:rotation}$, positive-semidefinitene of mass and inertia tensor $\eqref{opti:mass}$ and a triangular inequality$\eqref{opti:tri}$, where $J_{(\cdot)}$ denotes the eigenvalues of the inertia tensor of a rigid body ($\br{I}=\mathbf{R}\mathbf{J}\mathbf{R^T}$). 
The solution to this problem is possible using nonlinear optimization on manifolds. In our case, we implicitly consider dynamic consistency by using the regularization (see Eq. \ref{nnloss}). 

\subsection{Real-to-Sim Adaptation}
In terms of dynamics, the \textit{reality gap} $\delta$ is caused by a combination of inaccurate parameters in the parametric model and non-parametric uncertainties (e.g., backlash, hysteresis, etc). The \textit{reality gap} $\delta$ regarding the nonlinear dynamics of a multi-body system can be decomposed into a parametric modeling error $\delta \hat{f}$ and non-parametric modeling error $\delta \hat{g}$ as follows:
\begin{equation}
\delta_t = \underbrace{\delta \hat{f}(\br{\hat{x}_t}, \br{\hat{u}_t} ; \boldsymbol{\zeta})}_{\text{Parametric Error}} + \underbrace{\delta \hat{g}(\br{\hat{x}_t}, \br{\hat{u}_t})}_{\text{Non-Parametric Error}} 
\end{equation}
\noindent where the system parameters  \( \boldsymbol{\zeta} \), the estimated states \( \br{\hat{x}} \) , the control effort \( \br{\hat{u}} \), and time \( t \). To compensate for the error caused by $\delta \hat{f}$ and $\delta \hat{g}$, we leverage the SysID and GPs, respectively. 

In this work, we handle the errors $\delta \hat{f}$ and $\delta \hat{g}$ separately, ensuring that each error is corrected in its own step rather than optimizing both simultaneously. Note that this approach focuses on reducing the \textit{reality gap} for a robot itself without considering object dynamics, eliminating the need for real object datasets during training, but still leaves a \textit{reality gap} in object dynamics.

%% file: 4_method.tex
\section{METHODOLOGY}
\label{M:1}

Our framework consists of two distinct steps: 1) Real-to-Sim Adaptation aimed at enhancing the fidelity of the simulation via \textit{Robot System Identification} (SysID) and \textit{Gaussian Processes} (GPs); and 2) Learning the inertial parameters for an unknown object via time-series data-driven regression model. As a preliminary step for developing the end-to-end inertial parameter estimation using a high-fidelity simulation, we assume that the object can be perfectly held by the robotic hand, and a single trajectory is leveraged \cite{wu2012semi}. The overview of the proposed method is described in Algorithm \ref{alg:cap}.


\begin{figure}[t]
\begin{center}
\includegraphics[width=0.6\linewidth]{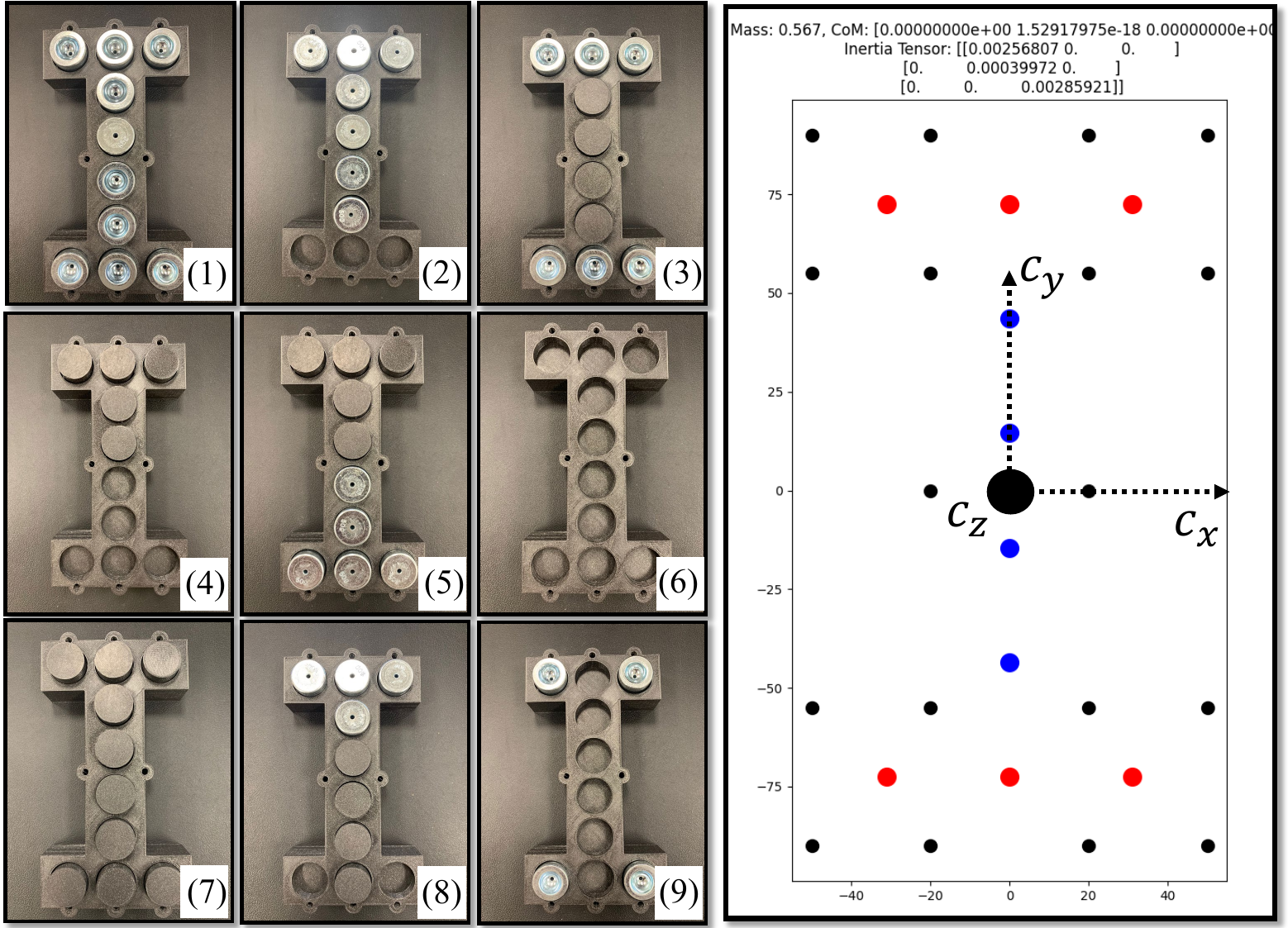}
\end{center}
\caption{\textbf{Customized Object To Get Ground-Truth Inertial Parameters.} Depending on the location of the weights, the object can be represented as a barbell(3), hammer(8), etc. Right side image shows a sample of an object(3) with ground truth inertial parameters. The CoM is defined based on a central fixed coordinate system.}
\label{fig2}
\vspace{-1.5em}
\end{figure}

\subsection{Target System Description and Unknown Object Design}
In this work, we utilize a four degrees-of-freedom (DoF) manipulator of a wheeled humanoid, SATYRR \cite{purushottam2022hands}, to hold and shake an object. To focus on manipulation part, we fix its torso to minimize the effect of the dynamics of whole-body SATYRR. Inspired by the previous work \cite{nadeau2022fast}, our object is designed to easily and precisely calculate the ground truth inertial parameter of various objects. This object consists of a combination of three cuboids, ten-cylinder holes, and steel weights. These weights can be placed in ten different locations as depicted in Fig.\ref{fig2}. The ground truth of inertial parameters can be calculated based on the moment of inertia, the density of each weight, and the size of each cuboid.

\subsection{Real-to-Sim Adaptation}
\subsubsection{Offline Robot System Identification} 
The goal of SysID is to minimize the discrepancy between a source data $\br{X_S}$ and a target data $\br{X_T}$ by searching for more realistic robot's system parameters $\boldsymbol{\zeta}$ in a simulation. The cost function can be some distance function $d(\cdot,\cdot)$ over a dataset of $m$ samples where $T$ is the length of each sample (Eq. (\ref{costfunc})). We chose the mean square error (MSE) as a $d$ function that is commonly used in SysID \cite{wensing2017linear}.
\begin{equation}
\mathcal{L}=\argmin_{\boldsymbol{\zeta}} \dfrac{1}{m}\sum_{i=1}^{m}\sum_{t=1}^{T}d(\br{X^t_S},\br{X^t_T}).
\label{costfunc}
\end{equation}
The optimal parameter $\boldsymbol{\zeta}^*$ can be chosen by solving the Eq. \ref{costfunc} and we utilized the particle swarm optimization (PSO) algorithm due to its global search ability and fast convergence speed. In this work, data $\br{X} \in \R^{8}$ contains position $q$ and velocity $\dot{q}$ in four joints of the manipulator in SATYRR. The parameters $\boldsymbol{\zeta} \in \R^{8}$ to be optimized contain joint damping 

\begin{algorithm}[H]
    \caption{Procedure of Learning Inertial Parameter Estimation of unknown Object via Real-to-Sim Adaptation}\label{alg:cap}
    \begin{algorithmic}
    \State \textit{\textbf{Phase 1}} \textit{Robot System Identification}
    \State \textbf{Input:} Target data $\mathcal{D_T}$ (Free case - not holding an object)
    
    \State \textbf{Initialize:} $\boldsymbol{\zeta} \in\mathbb{R}^8$, $\mathcal{L}\in \mathbb{R}$ 
    \While{until $ (\mathcal{L} \text{  converges})$}  
    \State $\br{X_T}= \{ q^{1:4}, \dot{q}^{1:4}\} \gets \mathcal{D_T} $
    \State $\br{X_S}= \{ \hat{q}^{1:4}, \hat{\dot{q}}^{1:4}\} \gets \hat{f}(\cdot;\boldsymbol{\zeta})_{sim}$ \hfill\Comment{IssacGym}
    
    \State Calculate the Loss function $\mathcal{L}$ \hfill\Comment{Eq. (\ref{costfunc})}
    \State Update parameter $\boldsymbol{\zeta}$ via \textit{particle swarm optimization} 
    \State \textbf{Return:} $\boldsymbol{\zeta^*}$
    \EndWhile
    \vspace{0.5em}  
    
    \State \textit{\textbf{Phase 2}} \textit{Gaussian Processes} 
    \State \textbf{Input:} \textit{Gaussian Processes} $\mathcal{GP}$, Simulation with SysID applied $\hat{f}(\cdot;\boldsymbol{\zeta^*})_{sim}$, and $\mathcal{D_T}$
    \State \textbf{do}
    \State \hspace{1em} $\br{X_T}= \{ q^{1:4}, \dot{q}^{1:4}\} \gets \mathcal{D_T} $
    \State \hspace{1em} $\br{X_S}= \{ \hat{q}^{1:4}, \hat{\dot{q}}^{1:4}\} \gets \hat{f}(\cdot;\boldsymbol{\zeta^*})_{sim}$
    \State \hspace{1em} Calculate non-parametric modeling error $\hat{g}(\br{\hat{x}}, \br{\hat{u}},t)$
    \State \hspace{1em} $(\hat{g}(\br{\hat{x}}, \br{\hat{u}},t) = \br{X_T} - \br{X_S})$
    \State \hspace{1em} Optimize a GP regression model using GPy. \hfill\Comment{Eq. (\ref{gpeqn})}
    \State \hspace{1em} Save a GP model $\mathcal{GP}(\mu, k)$
    \State \hspace{1em} Apply the $\mathcal{GP}(\mu, k)$ to the simulation $\hat{f}(\cdot;\boldsymbol{\zeta^*})_{sim}$
    \State \hspace{1em} \textbf{Return:} $\hat{f}(\cdot;\boldsymbol{\zeta^*},\mathcal{GP})_{sim}$
    \State \textbf{end}
    
    \vspace{0.5em}
    \State \textit{\textbf{Phase 3}} Learning Inertial Parameter of an Object
    \State \textbf{Input:} $\mathcal{D_S} \gets \hat{f}(\cdot;\boldsymbol{\zeta^*},\mathcal{GP})_{sim}, \mathcal{L_{NN}}$ 
    \While{until $ (\mathcal{L_{NN}} \text{  converges})$} 
    \State \hspace{1em} Training a time-series data-driven model \ref{sec::trainingnn}
    \State \hspace{1em} to minimize the loss $\mathcal{L_{NN}}$  \hfill\Comment{Eq. (\ref{nnloss})}
    \EndWhile
    \end{algorithmic}
\end{algorithm}

\noindent $\br{d_j} \in \R^{4}$ and link mass $\br{m_l} \in \R^{4}$. These parameters are chosen based on the major components that impact the dynamics. The center of mass and inertia tensor are determined using the CAD file, after setting the material and the coordinate center. To create a realistic simulation environment, we set the PD controller gains ($\br{K_p} \in \R^{4}, \br{K_d} \in \R^{4}$) based on the real-world gains used for SATYRR, with a control frequency of 400Hz for both simulation and real-world scenarios.

\subsubsection{Non-Parametric Dynamics Modeling via \textit{Gaussian Processes}}
\begin{figure}[t]
     \centering
 	\begin{subfigure}[b]{0.4\linewidth}
 		\includegraphics[width=0.95\columnwidth]{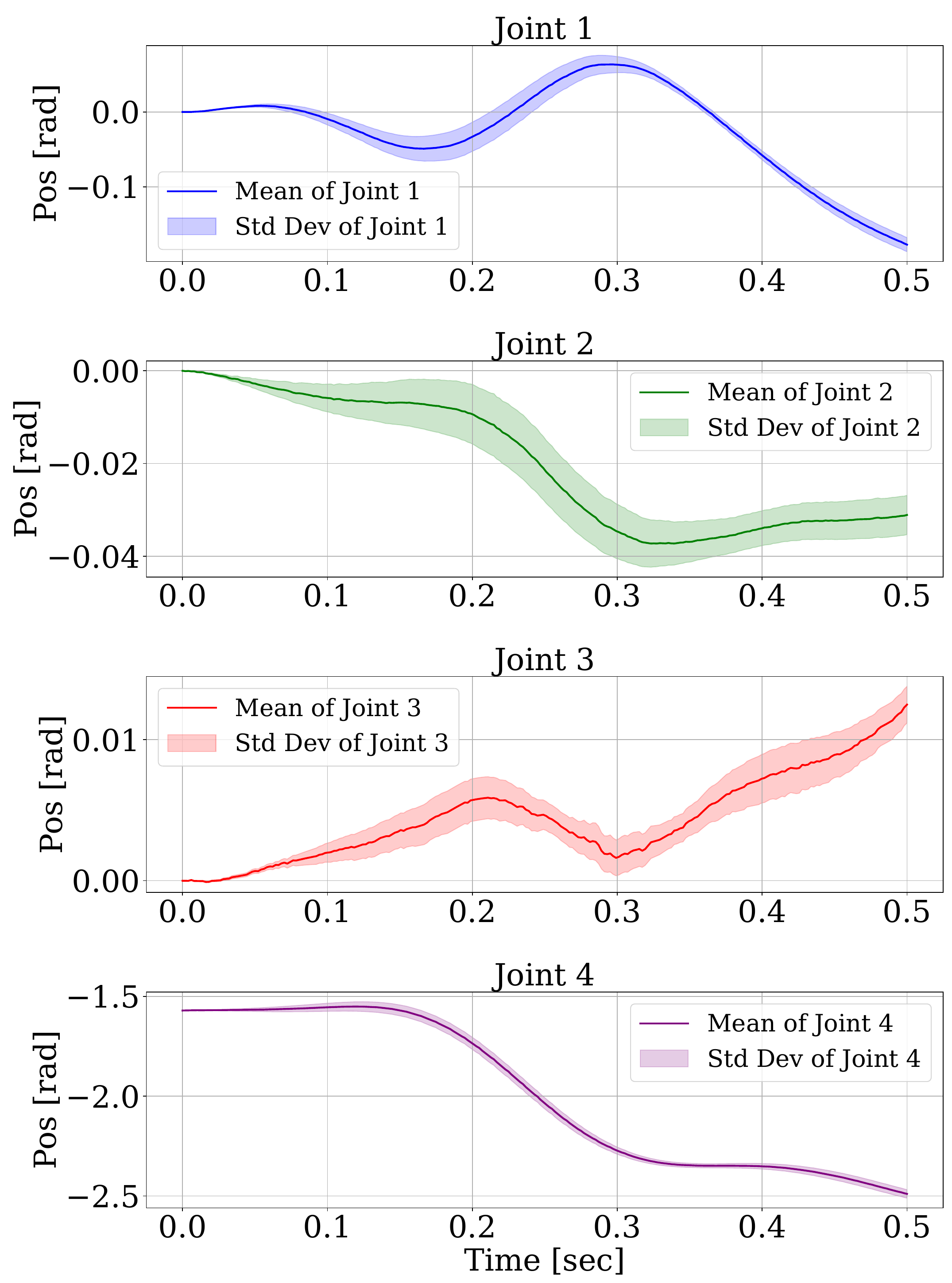}
            \caption{Joint-level Position}
 	\end{subfigure} 
 	\begin{subfigure}[b]{0.4\linewidth}
 		\includegraphics[width=0.95\columnwidth]{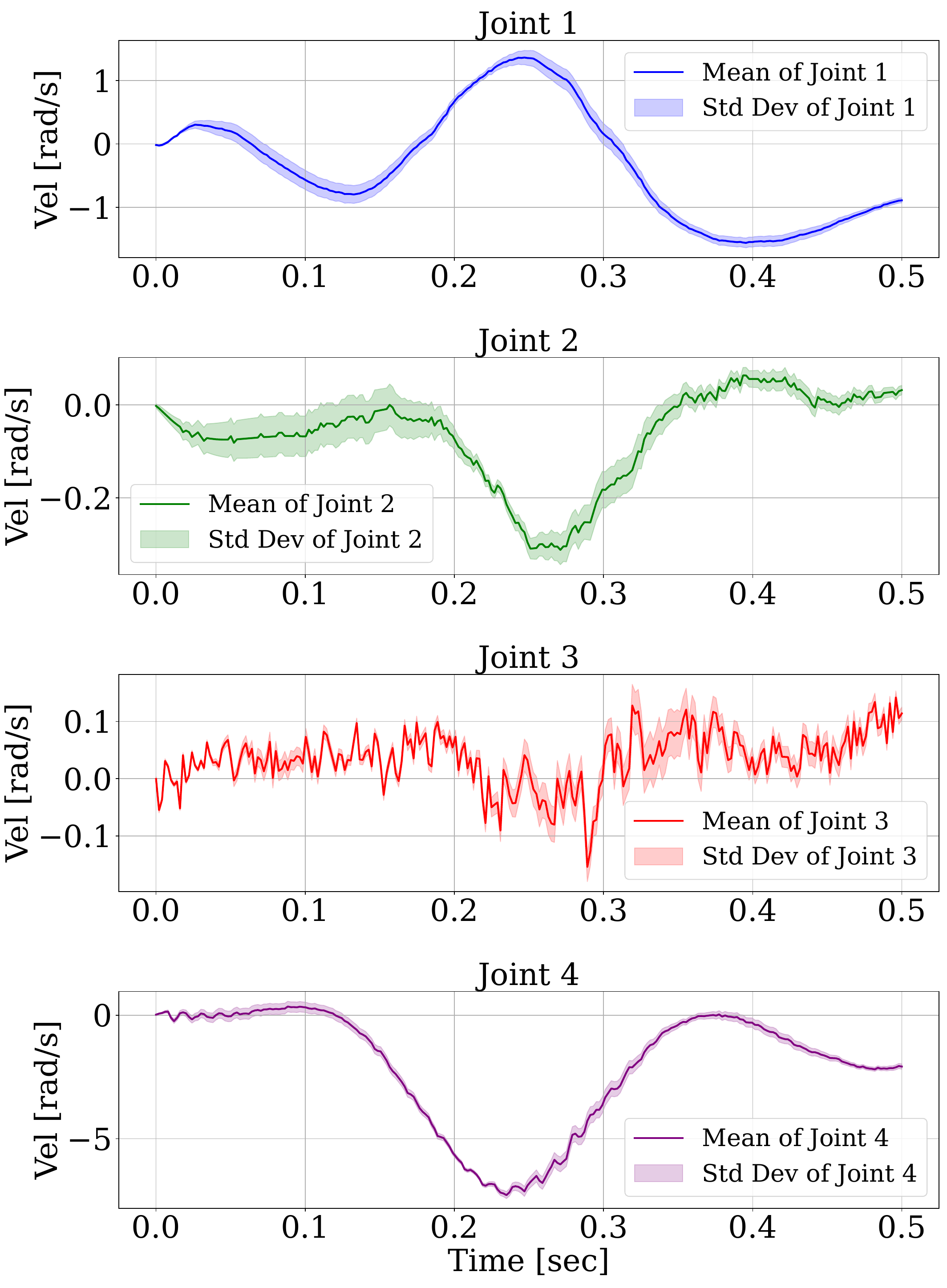}
 		\caption{Joint-level Velocity}
 	\end{subfigure} 
        \caption{\textbf{Input Data Distribution in Training Dataset}. The graphs shows the mean and standard deviation of joint trajectories in training dataset. This represents how different object's dynamics properties affect the trajectories of joint position and velocity for each joint. }
 	\label{sim2real_fig}
\vspace{-1.5em}
\end{figure}

To model the reality gap caused by non-parametric modeling error $\delta \hat{g}(\br{\hat{x}_t}, \br{\hat{u}_t})$, such as nonlinear friction and backlash, we leverage Gaussian Processes (GPs). GPs offer the benefit of sample efficiency and the potential to construct a wide range of functions without assuming a specific functional form \cite{rasmussen2006gaussian}. Given the different scales between joint position and velocity, we employed separate GP regression models to accurately depict the residual errors for joint position and velocity, respectively. (e.g., output of GPs $y = \br{X_T} - \br{X_S}$ and $\br{X_S}$ is acquired from a simulation $\hat{f}(\cdot;\boldsymbol{\zeta}^*)_{sim}$ which utilizes the optimal parameters $\boldsymbol{\zeta}^*$ from SysID). The non-parametric dynamics model $\hat{g}$ can be formulated with GPs as follows:
\begin{equation}
\hat{g}(\br{\hat{x}_t}, \br{\hat{u}_t}) = \mathcal{GP}(\mu(\br{\hat{x}_t}, \br{\hat{u}_t}), k(\br{\hat{x}_t}, \br{\hat{u}_t}, \br{\hat{x}_{t'}}', \br{\hat{u}_{t'}}')).
\label{gpeqn}
\end{equation}
here, the individual terms represent the state \( \br{\hat{x}} \), control effort \( \br{\hat{u}} \), and time \( t \). The GPs \( \mathcal{GP}(\mu, k) \) is expressed with a mean function \( \mu \) and covariance function \( k \) where the mean \( \mu(\br{\hat{x}_t}, \br{\hat{u}_t}) \) represents the expected value of the process at each point in the input space and the covariance function \( k(\br{\hat{x}_t}, \br{\hat{u}_t}, \br{\hat{x}_{t'}}', \br{\hat{u}_{t'}}') \) describes the correlation between outputs for two distinct sets of inputs in the function \( \hat{g} \). We constructed and optimized the GP regression model using the GPy library. Radial Basis Function (RBF) kernel (both variance and length scale set to 1 for position and 50 for velocity) is utilized. The RBF kernel parameters were manually selected until the \textit{reality gap} was sufficiently reduced (see Fig. \ref{sim2real_fig}). 

\subsection{Learning Inertial Parameter of Unknown Object}
\label{C_METHOD}

\subsubsection{Dynamic Trajectory Planning and Manipulation Control}
Periodic excitation trajectories, often based on Fourier series and trigonometric functions, are commonly used for dynamic model identification \cite{wensing2017linear,nadeau2022fast,wang2019convex}. While effective, they can be time-consuming (e.g., 35 seconds in \cite{nadeau2022fast}). Inspired by how humans shake objects to identify them (with decreasing amplitude and frequency over time), we designed the dynamic trajectory as follows.
\begin{gather}
  f(t) = h_{\text{freq}} - (h_{\text{freq}} - l_{\text{freq}}) t/T, \hspace{0.5cm} \phi(T) = \Delta t \sum_{\tau=0}^{T/\Delta t} f(\tau) \\
  w(t) = \frac{1}{2} \left(1 - \cos\left(\frac{2 \pi t}{T}\right)\right) \\
  X(t) = \alpha \sin(2 \pi \phi(T)) w(t)
\label{destraj}
\end{gather}
\noindent where the frequency \( f(t) \) starts at \( h_{\text{freq}} \) and linearly decreases to \( l_{\text{freq}} \). The signal phase, \( \phi \), is computed as the cumulative sum of the frequency, scaled by the time step. The sinusoidal trajectory \( X(t) \) has an amplitude scaled by a factor \( \alpha \) and is modulated by a \textit{Hann window}, \( w(t) \), for smooth transitions. The trajectory $X(t)$ is applied to the end-effector in $x$ and $y$ axis  ($T=0.5$, x-axis: $h_{freq}=5, l_{freq}=1, \alpha=-1$, y-axis: $h_{freq}=3, l_{freq}=1, \alpha=5$). The numerical inverse kinematics using a pseudo-inverse Jacobian of the 4-DoF manipulator is employed to control the manipulator of SATYYR, which is defined as
\begin{equation}
\begin{aligned}
J^+ & = (J^T J + \lambda^2 I)^{-1} J^T, \\
\theta_{des} & = \theta + \lambda  J^+ e.
\end{aligned}
\label{trajectoryeqn}
\end{equation}
where $J^+$ represents a pseudo-inverse Jacobian leveraging a \textit{Damped Least Squares method}, aiding to improve stability near a singular configuration and dealing with noisy data or slight modeling errors by utilizing regularization. The symbol $\lambda$ is damping, $\alpha$ is a hyperparameter, and $e = x_{des} - x$ is task-space position error.

\subsubsection{Database Construction}
To build a learning-based object dynamics estimator, we constructed the $M$ number of source dataset $\mathcal{D}_S = { (\br{X}^i_S,\br{y}^i_S)}^{M}_{i=1} (M=5000)$ using the IsaacGym simulator \cite{makoviychuk2021isaac}, with 5,000 environments running simultaneously. To directly apply the pre-trained estimation model to the SATYRR without further manual tuning, we employed a more realistic simulator to acquire the dataset $\mathcal{D}_S$. All agents in the simulator track the desired trajectory (Eq. \ref{destraj}) while holding different objects. The estimator takes as input the concatenated vector $\br{X_S}$ and outputs estimated inertial parameter $\br{y_S}$ as:
\begin{equation}
\begin{aligned}
& \br{X_S}= [q^1_{t:T}, q^2_{t:T}, q^3_{t:T}, q^4_{t:T}, \dot{q}^1_{t:T}, \dot{q}^2_{t:T}, \dot{q}^3_{t:T}, \dot{q}^4_{t:T}]^\top \\
& \br{y_S}= [m, c_x, c_y, c_z, I_{xx}, I_{yy}, I_{zz}]^\top 
\end{aligned}
\label{inputandoutput}
\end{equation}
where $q^i_{t: T}$ and $\dot{q}^i_{t: T}$ are the history of joint position and velocity of each joint, respectively. The term $\br{y_S}$ denotes the vector of inertial parameters. Each parameter component is either 1) randomly generated with a uniform distribution, or 2) determined by adjusting the weight location, length scale (width and height of each cuboid), and density. After random generation, we verified their mass and ensured the inertia tensor satisfies the triangular inequality for physical feasibility. We simplify the problem by focusing on estimating the diagonal terms in the inertia matrix considering a more tractable analysis while retaining the essential physics of the problem. To collect real-world data for validation, the wheeled humanoid robot SATYRR \cite{purushottam2022hands} tracked a pre-recorded shaking motion generated in simulation (see Eq. (\ref{trajectoryeqn})). A total of ten different objects were involved, along with a free case where no object was held. For each object case, the trajectory was applied to the SATYRR five times (i.e., a total of 45 real-world dataset were obtained for the validation in the experiment).

\subsubsection{Training Time-Series Data-Driven Model with Dynamic Consistency Regularization}
\label{sec::trainingnn} To learn the inertial parameter $\br{\phi}$, we trained a time-series data-driven regression model (e.g., 1D-CNN) using 4000 samples of the training set in dataset $\mathcal{D}_S$ (500 for validation set and 500 for test set). Dynamic consistency is implicitly taken into account by applying the regularization term in the loss function to estimate more physically reliable parameters. The total loss function considering dynamic consistency is defined as    
\begin{equation}
\begin{aligned}
& \mathcal{L_{NN}} = w_1 \frac{1}{n}\sum_{i=1}^{n}(\br{y^i_S} - \br{\hat{y}^i_S})^2 + w_2 \mathcal{L_{\text{tri}}} +w_3\mathcal{L_{\text{pos}}}\br{\hat{y}_S} \\ 
& \mathcal{L_{\text{tri}}}(I) = \frac{1}{n} \sum_{i=1}^{n} \sum_{j=1}^{3} \text{ReLU}\left( I_{j}^i - \sum_{k=1}^{3} I_{k}^i \right) \\
& \mathcal{L_{\text{pos}}}\hat{y} = \frac{1}{n} \sum_{i=1}^{n} \left| \min \left( \br{\hat{y}^i_S}(k), 0 \right) \right|, \ k=[0, 4, 5, 6]
\end{aligned}
\label{nnloss}
\end{equation}

\begin{figure*}[t]
     \centering
 	\begin{subfigure}[b]{0.24\linewidth}
 		\includegraphics[width=\columnwidth]{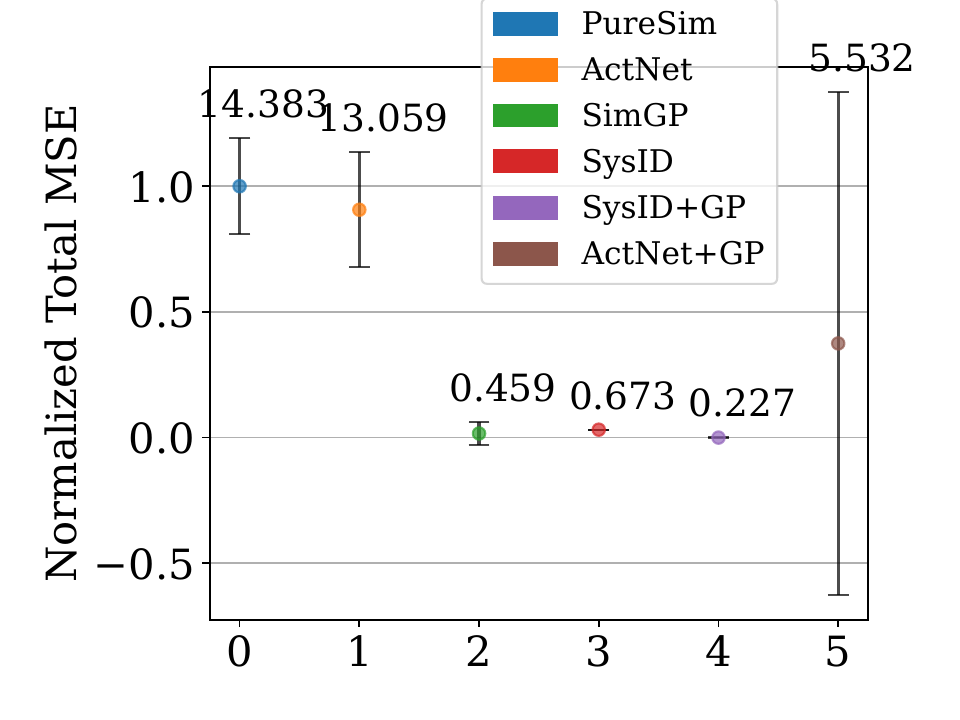}
 		\caption{Total: Position+Velocity}
 	\end{subfigure} 
 	\begin{subfigure}[b]{0.24\linewidth}
 		\includegraphics[width=\columnwidth]{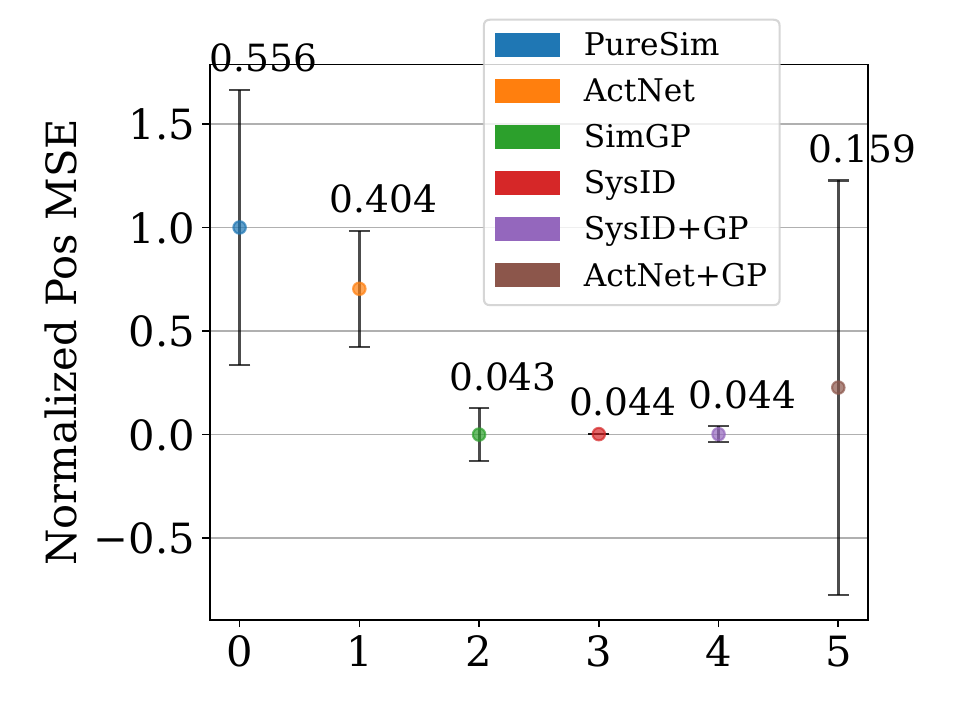}
 		\caption{Position}
 	\end{subfigure} 
        \begin{subfigure}[b]{0.24\linewidth}
 		\includegraphics[width=\columnwidth]{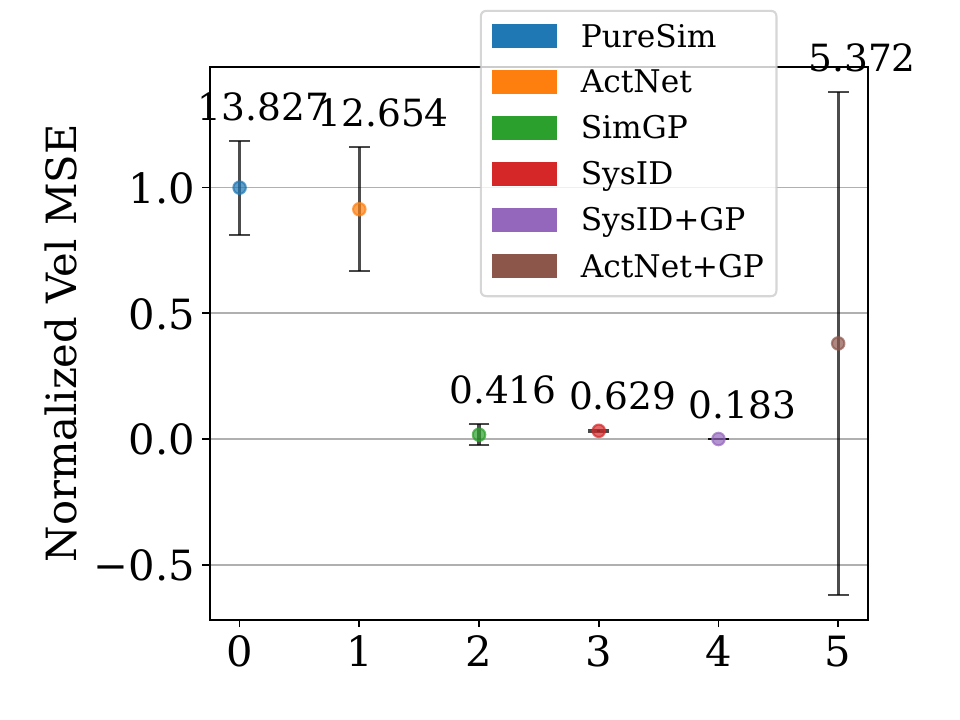}
 		\caption{Velocity}
 	\end{subfigure} 
        \caption{\textbf{Results of Real2Sim Adaptation}. The graphs depict the mean and standard deviation of the normalized Mean Squared Error (MSE) between trajectories derived from simulation and the real world, employing min-max normalization for straightforward comparison. The 45 target objects, which are not considered in the optimization process, are utilized for evaluation. The numbers on the graph represent the MSE. Based on the normalized total MSE outcomes, \textbf{Sim+SysID+GP} exhibits the smallest \textit{reality gap}. The MSE for \textbf{Sim+SysID} is also notably low, showing only a minor performance gap compared to \textbf{Sim+SysID+GP}. This is because the error of parametric model has a large portion in causing the \textit{reality gap}. The error from non-parametric modeling can become substantially larger in scenarios involving contact. }
 	\label{sim2real_fig2}
\vspace{-0.5em}
\end{figure*}

\noindent where the total loss $\mathcal{L_{NN}}$ combines mean square error, triangular inequality loss $\mathcal{L_{\text{tri}}}(I^{(\cdot)})$, and negative output penalization loss $\mathcal{L_{\text{pos}}}\br{\hat{y}_S}$. The term $w_1, w_2$, and $w_3$ are the weight of each term and are decided by manual tuning considering the importance of each parameter. We have empirically observed that neural network is capable of learning inertial parameters without accounting for the varying scales of these parameters. Our 1D CNN architecture processes sequential data with four 64-filter convolutional layers (kernel size 3, stride 1, padding 1), ReLU activations, and a MaxPooling layer (kernel size 2). It concludes with a dense network transitioning from 512 to 128 neurons, leading to the specified output size.



%% file: 5_experiment.tex
\section{EXPERIMENT}
\label{E:1}

The proposed learning-based inertial parameter identification method is verified with a 4-DoF manipulator of SATYRR \cite{purushottam2022hands} in both a simulation and the real world. The experimental setup is illustrated in Fig. \ref{fig1} and Fig. \ref{fig2}. Pre-defined trajectory, SATYRR robot, HMI, and ground-truth of inertial parameters are utilized. We conducted three major experiments in total. Details are described below. 

\subsection{Real-to-Sim Adaptation} The main purpose is to validate our sim-to-real adaptation by answering two questions: (1) How effectively can SysID and GPs compensate for parametric and non-parametric modeling errors to reduce the reality gap? and (2) Does this method generalize to test environments where object dynamics are taken into account? We evaluate the real-to-sim adaptation ability by comparing six separate baselines: (a) \textbf{PureSim}: simulation using default physics parameters and PD controller with almost perfect tracking performance (b) \textbf{Sim+ActNet}: using Actuator Network  \cite{hwangbo2019learning} instead of PD controller. We collected a dataset from SAYTRR and trained the Actuator Network parameterized with $\theta$ defined as 
\begin{equation}
    \tau_j = h_{\theta}(q^*_j, q_j, \dot{q}^*_j, \dot{q}_j, K_p, K_d), \hspace{0.5cm} j=1,2,3,4
    \label{}
\end{equation}
where desired joint position and velocity $q^*$ and $\dot{q}^*$, actual joint position and velocity $q$ and $\dot{q}$, and controller gain $K_p$ and $K_d$.
(c) \textbf{Sim+GPs}: applying the GPs into PureSim (d) \textbf{Sim+SysID}: applying the SysID into PureSim (e) \textbf{Sim+SysID+GPs}: Applying the GPs to (d) (f) \textbf{ActNet+Gps}: applying the GPs to (b). The mean sqaure error (MSE, Eq. \ref{costfunc}) is used as an evaluation index. Note that the GP model is trained exclusively on data without objects, yet it's tested under conditions where the robot interacts with objects. This tests the adaptability of the optimized SysID and GPs to unseen dynamics, evaluating their capability to accurately estimate object inertial parameters in new scenarios.

\subsection{Inertial Parameter Estimation}
This experiment addresses the following questions: (1) How fast and accurate is our framework compared to previous methods for estimating the inertial parameters of unknown objects? Which neural network models are most suitable? (2) How does our method perform in real-world scenarios, and does our adaptation method improve estimation performance? (3) How does our method compare to previous works, and can it be applied to a more generalizable object dataset? We evaluated the performance of our framework against the following baselines: (1) Ordinary Least Squares (OLS) \cite{atkeson1986estimation} and (2) Weighted Least Squares (WLS) \cite{wang2019convex}. Since these traditional methods require a force/torque sensor, we used 45 samples from the simulation test dataset to directly assess force and torque values. For a fair comparison, we used the open-source MATLAB code provided by the authors \cite{atkeson1986estimation,wang2019convex}. We also adopted TuneNet \cite{allevato2020tunenet}, DROPO \cite{tiboni2023dropo} and OSI \cite{yu2017preparing} as learning-based baseline methods. The mean absolute error (MAE=$|\hat{y}_k - y_k|$) and the normalized MAE (NMAE= $\frac{1}{n} \sum_{k=1}^{n} \frac{| \hat{y}_k - y_k |}{y_{\text{scale}}}$
) are leveraged as evaluation indexes which are commonly chosen in this field \cite{lao2023learning,sundaralingam2021hand}.

\subsection{Control Task Experiments}
How beneficial are the identified inertial parameters for improving model-based controller performance, even with imperfect estimation accuracy? To verify this, we conducted real-world manipulation and locomotion tasks using a SATYRR robot holding a 1.5 kg drill (see Fig. \ref{task_fig}). In the manipulation task, we assessed the robot's trajectory tracking with the object, which is essential for warehouse tasks. For locomotion, we evaluated the robot's proficiency in transporting the object.

\begin{figure}[h]
     \centering
 	\begin{subfigure}[b]{0.48\linewidth}
 		\includegraphics[width=\columnwidth]{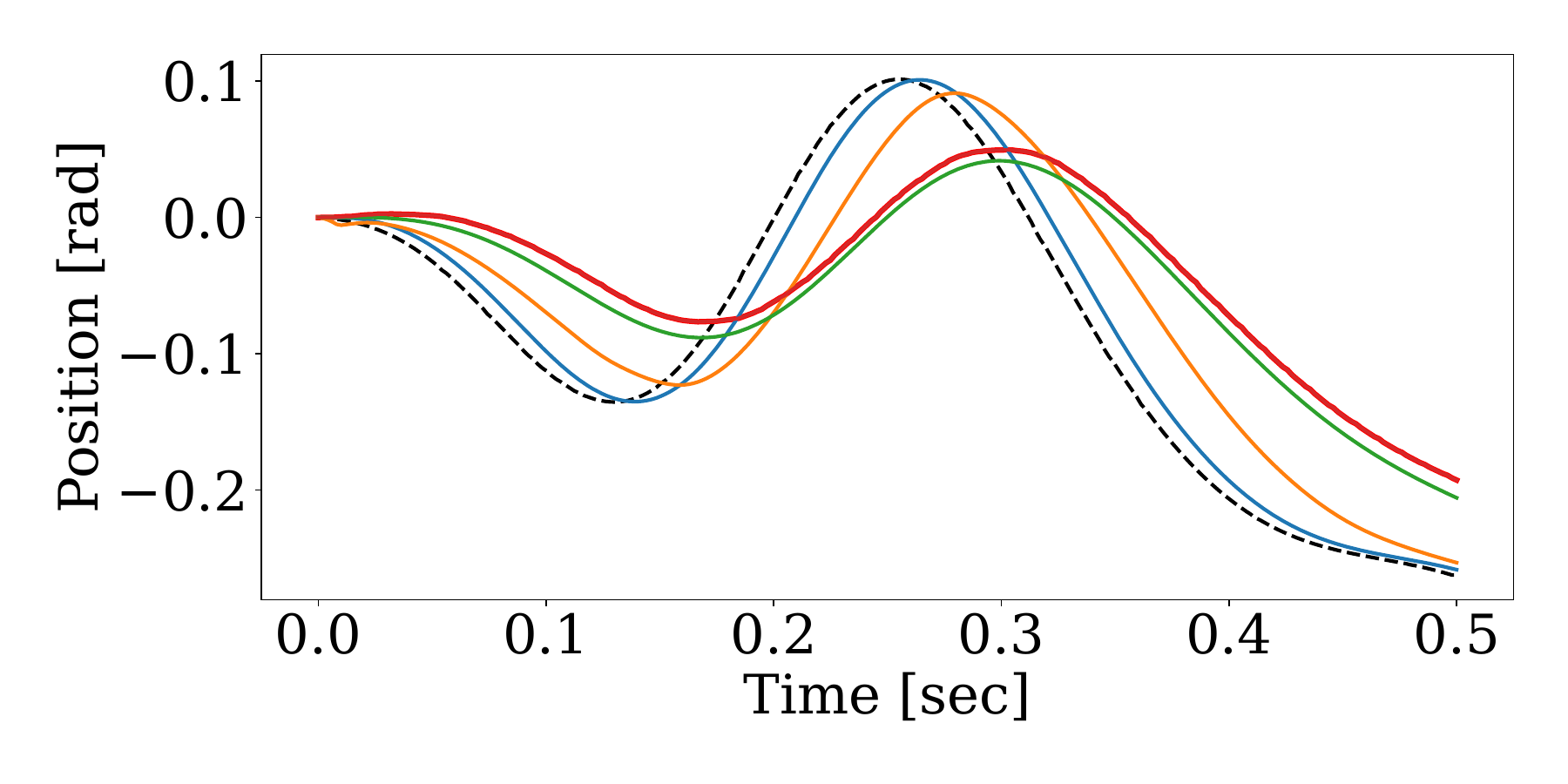}
 		\caption{Joint 1 Position}
 	\end{subfigure} 
 	\begin{subfigure}[b]{0.48\linewidth}
 		\includegraphics[width=\columnwidth]{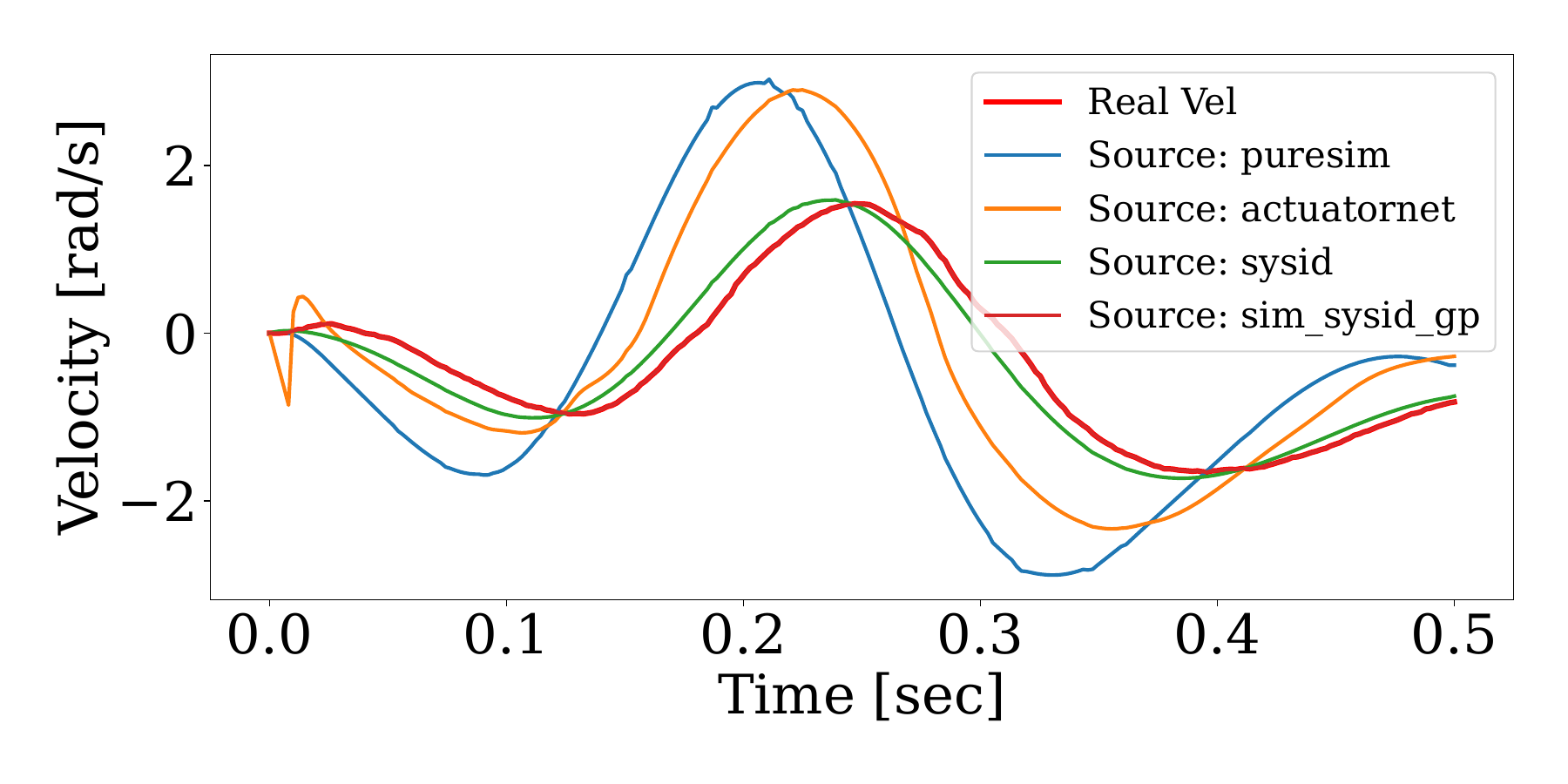}
 		\caption{Joint 1 Velocity}
 	\end{subfigure} 
 	\caption{\textbf{Trajectory Comparison Results Obtained From Real World and Simulation}. The trajectory produced by \textbf{Puresim} is closely located with the desired trajectory (black). First, \textbf{SysID} is applied to \textbf{PureSim}, followed by \textbf{GP} to \textbf{SysID} to handle the errors in parametric and non-parametric modeling, respectively. Each step brings the simulation trajectory closer to the target trajectory (red). In the case of \textbf{Sim+SysID+GP}, all joint trajectories in position and velocity are almost perfectly matched with the actual trajectories.}
\label{sim2real_fig}
\vspace{-1.5em}
\end{figure}

%% file: 6_results_and_discussion.tex
\section{RESULTS AND DISCUSSION}
\label{Results}

\subsection{Real-to-Sim Adaptation} As shown in the Fig \ref{sim2real_fig}, the SysID and GPs contribute to reduce the \textit{reality gap} in both position and velocity trajectories. In the case of GPs, it learns not only non-parametric error but also noise in real data.

Although we successfully reduced the \textit{reality gap} in the robot system, residual dynamics errors caused by the object's dynamics still remain. The results illustrated in Figure \ref{sim2real_fig2} shows that our method, developed without object dynamics information, performs effectively even in a situation where new object is applied. Implementing the SysID on a \textbf{PureSim} environment significantly diminishes the \textit{reality gap} for all examined object cases. Note that none of objects are utilized in optimizing the SysID and GPs. Overall, \textbf{Sim+SysID+GP} exhibits the smallest \textit{reality gap} in comparison to other approaches. It has been observed that incorporating SysID contributes to a more generalizable adaptation from real to simulated environments. This is evidenced by the superior performance of \textbf{Sim+SysID+GP} over the use of \textbf{Sim+GP}, even though \textbf{Sim+GP} also perfectly reduce the reality gap within the training set. This supports that the parametric model can be more generalizable than non-parametric model using the GPs model. While the Actuator Network demonstrated potential in narrowing the \textit{reality gap} \cite{hwangbo2019learning}, it fell short in aligning the trajectories between the simulated and real environments. This discrepancy suggests that while the Actuator Network can help RL algorithms learn more realistic actions based on more accurate state information, it does not guarantee a complete convergence of trajectories across the two domains. The similar result can be observed in \cite{li2023robust}.


While the \textbf{Sim+SysID+GP} approach has shown effectiveness in closing the reality gap across different objects, it still requires enhancements in several aspects. The real-to-sim adaptation relies on a single trajectory, making its effectiveness highly dependent on the selected trajectory. This constraint is particularly pronounced due to the reliance on GPs \cite{wu2012semi} and the trajectory used for SysID. It is well-known that the performance of data-driven models is significantly influenced by the training data distribution. Exploring and combining physics-based and data-driven models could be an interesting direction for future research.


\subsection{Inertial Parameter Estimation}
\subsubsection{Benchmark Results with Simulation Dataset}
\label{Benchmark Results with Simulation Dataset}
\begin{figure*}[t]
     \centering
 	\begin{subfigure}[b]{0.161\linewidth}
 		\includegraphics[width=\columnwidth]{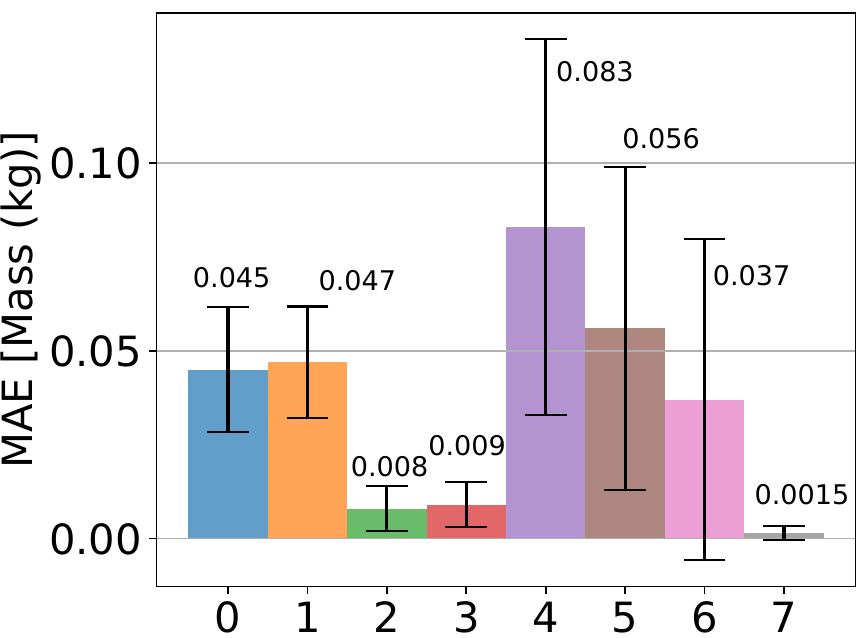}  
 	\end{subfigure}
        \begin{subfigure}[b]{0.161\linewidth}
 		\includegraphics[width=\columnwidth]{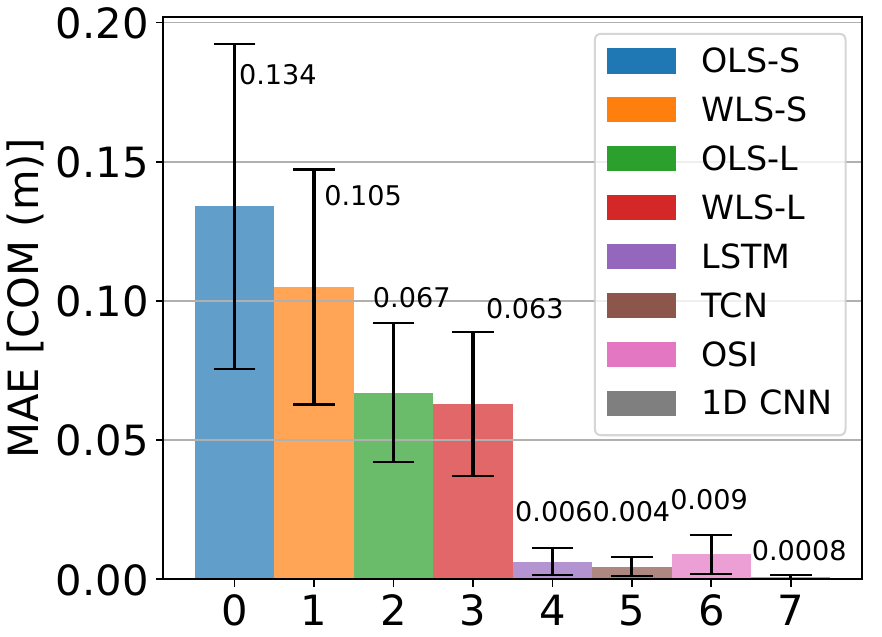}
 	\end{subfigure}
        \begin{subfigure}[b]{0.161\linewidth}
 		\includegraphics[width=\columnwidth]{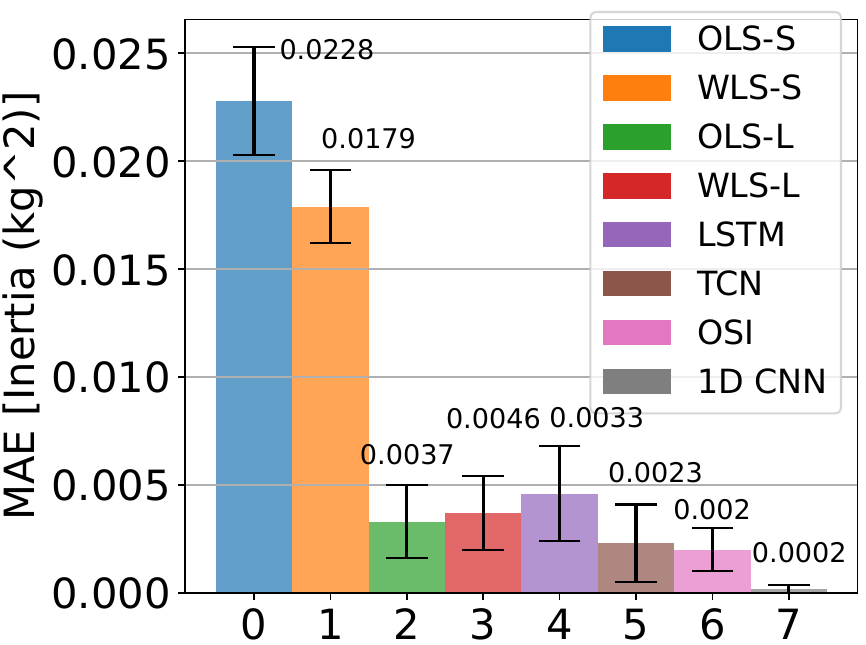}
 	\end{subfigure}
        \begin{subfigure}[b]{0.161\linewidth}
 		\includegraphics[width=\columnwidth]{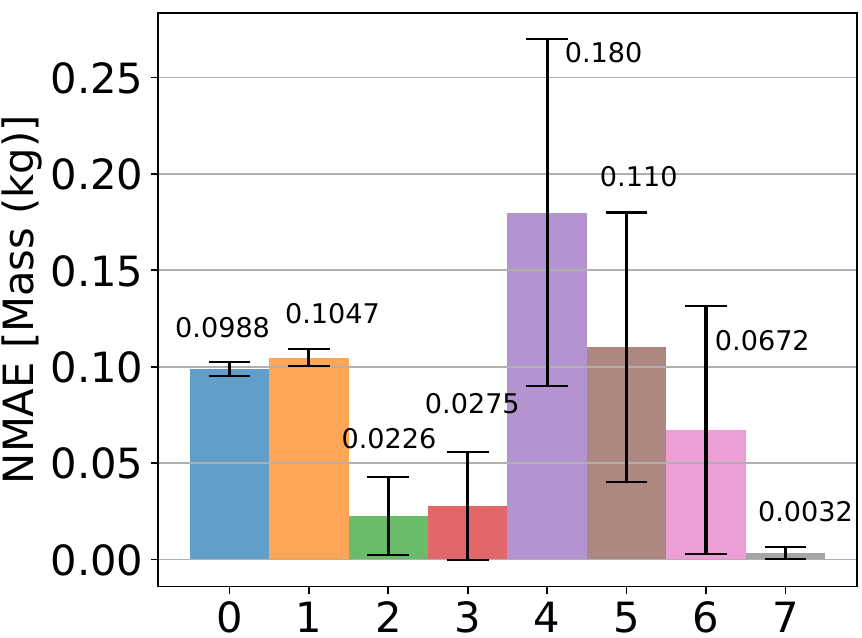} 
 	\end{subfigure} 
        \begin{subfigure}[b]{0.161\linewidth}
 		\includegraphics[width=\columnwidth]{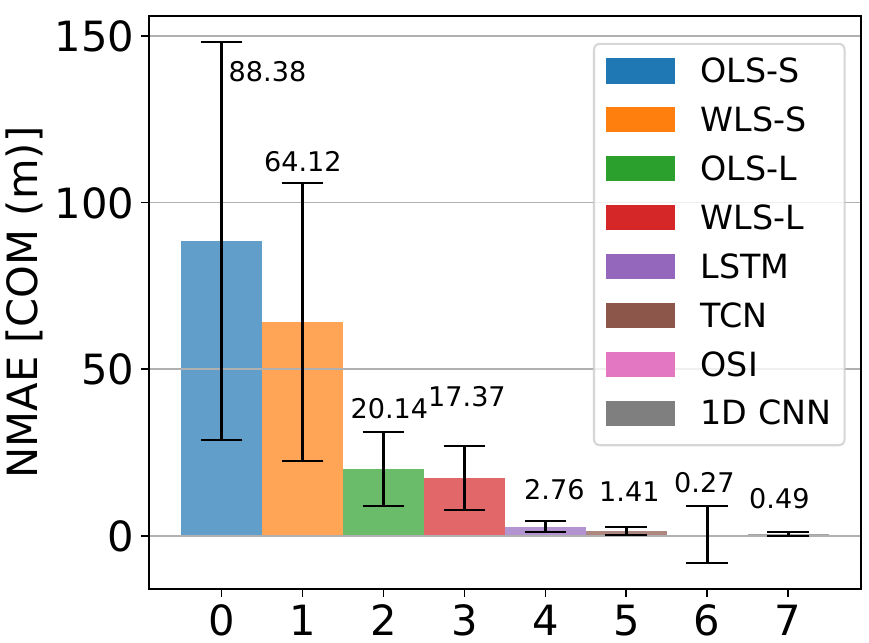}
 	\end{subfigure} 
        \begin{subfigure}[b]{0.161\linewidth}
 		\includegraphics[width=\columnwidth]{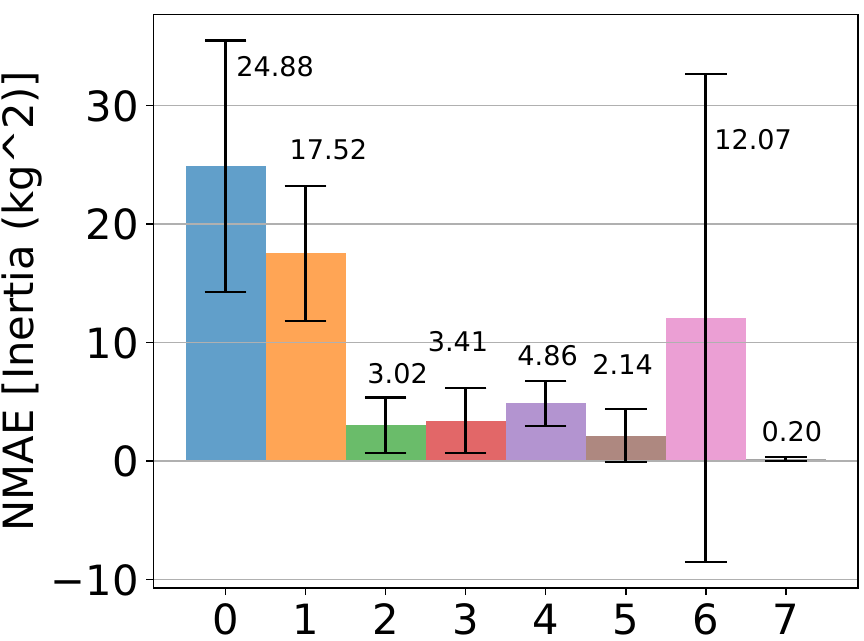}
 	\end{subfigure}
 	\caption{\textbf{Inertial Parameter Estimation Benchmark in Simulation}. Overall, 1D-CNN showed most accurate estimation performance compared to other baselines. The TuneNet exhibited poor performance in estimating mass, while demonstrating comparable performance in estimating the center of mass (CoM) and inertia. Learning-based methods generally outperformed conventional least-squares methods, highlighting their potential to achieve fast and accurate parameter estimation.}
\label{estimation_benchmark_sim}
\vspace{-1em}
\end{figure*}

We validated the proposed inertial parameter estimation framework against conventional methods in terms of accuracy, speed, and dynamic consistency (see Fig. \ref{estimation_benchmark_sim}). The simulation dataset is employed to obtain crucial measurements such as acceleration, torque, and force at the object level, which are fundamental elements in conventional methods. The proposed estimator using a 1D-CNN achieved the highest accuracy in estimating the inertial parameter of unknown objects with SATYRR. As shown in Table \ref{table3}, the total estimation time for our method is around 0.1 second including generating a trajectory (e.g., arm shaking motion). On the other hand, we observed that the existing methods such as OLS ans WLS took more time as they depend on a long persistently exciting signal, highlighting our method's potential for real-time object characteristic identification. The 1D-CNN surpassed LSTM and TCN in estimation performance, efficiently capturing local dependencies via Convolutional filters. This is because the distinct behaviors associated with dynamics are likely concentrated within specific segments of the entire trajectory. This localization makes the 1D-CNN particularly adept at identifying these critical features for enhanced performance. 

\begin{table*}[t]
\vspace{1em}
\centering
\caption{\textbf{Results of Inertial Parameter Estimation of Unknown Objects In the Real-World.} The numbers (e.g., A (B)) represent the mean (A) and standard deviation (B) of the estimation performance of our estimator (1D-CNN case) for nine different objects, as shown in Fig. \ref{fig2} of our manuscript. We excluded the results of the \textbf{PureSim}, \textbf{Sim+GP}, and \textbf{Sim+ActNet} models due to overfitting. The physically feasible value indicates the rate of non-feasible outcomes out of the total number of trials (zero means all outputs are physically feasible).}
\label{table1}
\setlength{\tabcolsep}{2.35pt}
\begin{tabular}{
>{\columncolor[HTML]{EFEFEF}}c |ccccccc|
>{\columncolor[HTML]{FFFFFF}}c |ccccccc|c|c}
\cline{1-8} \cline{10-16} \cline{18-18}
{\color[HTML]{000000} }                                                                                              & \multicolumn{7}{c|}{\cellcolor[HTML]{EFEFEF}{\color[HTML]{000000} Mean Absolute Error (MAE)}}                                                                                                                                                                                                                                                                                                                                                                                                                                                                                                                                                                                                                                                                                                                                                                                                                           & {\color[HTML]{000000} } & \multicolumn{7}{c|}{\cellcolor[HTML]{EFEFEF}{\color[HTML]{000000} Normalized Mean Absolute Error (NMAE)}}                                                                                                                                                                                                                                                                                                                                                                                                                                                                                                                                                                                                                                                                                                                                                                                                               &  & \cellcolor[HTML]{EFEFEF}{\color[HTML]{000000} }                                                            \\ \cline{1-8} \cline{10-16} \cline{18-18} 
\cellcolor[HTML]{EFEFEF}{\color[HTML]{000000} \textbf{\begin{tabular}[c]{@{}c@{}}Inertia\\ Parameters\end{tabular}}} & \multicolumn{1}{c|}{\cellcolor[HTML]{EFEFEF}{\color[HTML]{000000} \begin{tabular}[c]{@{}c@{}}Mass\\ (kg)\end{tabular}}} & \multicolumn{1}{c|}{\cellcolor[HTML]{EFEFEF}{\color[HTML]{000000} \begin{tabular}[c]{@{}c@{}}CoM x\\ (m)\end{tabular}}} & \multicolumn{1}{c|}{\cellcolor[HTML]{EFEFEF}{\color[HTML]{000000} \begin{tabular}[c]{@{}c@{}}CoM y\\ (m)\end{tabular}}} & \multicolumn{1}{c|}{\cellcolor[HTML]{EFEFEF}{\color[HTML]{000000} \begin{tabular}[c]{@{}c@{}}CoM z\\ (m)\end{tabular}}} & \multicolumn{1}{c|}{\cellcolor[HTML]{EFEFEF}{\color[HTML]{000000} \begin{tabular}[c]{@{}c@{}}Ixx\\ (kgm\textasciicircum{}2)\end{tabular}}} & \multicolumn{1}{c|}{\cellcolor[HTML]{EFEFEF}{\color[HTML]{000000} \begin{tabular}[c]{@{}c@{}}Iyy\\ (kgm\textasciicircum{}2)\end{tabular}}} & \cellcolor[HTML]{EFEFEF}{\color[HTML]{000000} \begin{tabular}[c]{@{}c@{}}Izz\\ (kgm\textasciicircum{}2)\end{tabular}} & {\color[HTML]{000000} } & \multicolumn{1}{c|}{\cellcolor[HTML]{EFEFEF}{\color[HTML]{000000} \begin{tabular}[c]{@{}c@{}}Mass\\ (kg)\end{tabular}}} & \multicolumn{1}{c|}{\cellcolor[HTML]{EFEFEF}{\color[HTML]{000000} \begin{tabular}[c]{@{}c@{}}CoM x\\ (m)\end{tabular}}} & \multicolumn{1}{c|}{\cellcolor[HTML]{EFEFEF}{\color[HTML]{000000} \begin{tabular}[c]{@{}c@{}}CoM y\\ (m)\end{tabular}}} & \multicolumn{1}{c|}{\cellcolor[HTML]{EFEFEF}{\color[HTML]{000000} \begin{tabular}[c]{@{}c@{}}CoM z\\ (m)\end{tabular}}} & \multicolumn{1}{c|}{\cellcolor[HTML]{EFEFEF}{\color[HTML]{000000} \begin{tabular}[c]{@{}c@{}}Ixx\\ (kgm\textasciicircum{}2)\end{tabular}}} & \multicolumn{1}{c|}{\cellcolor[HTML]{EFEFEF}{\color[HTML]{000000} \begin{tabular}[c]{@{}c@{}}Iyy\\ (kgm\textasciicircum{}2)\end{tabular}}} & \cellcolor[HTML]{EFEFEF}{\color[HTML]{000000} \begin{tabular}[c]{@{}c@{}}Izz\\ (kgm\textasciicircum{}2)\end{tabular}} &  & \cellcolor[HTML]{EFEFEF}{\color[HTML]{000000} \begin{tabular}[c]{@{}c@{}}Physically\\ Feasibility\end{tabular}} \\ \cline{1-8} \cline{10-16} \cline{18-18} 
{\color[HTML]{000000} \textbf{\begin{tabular}[c]{@{}c@{}}Sim\\ +ActNet+GP\end{tabular}}}                             & \multicolumn{1}{c|}{{\color[HTML]{000000} \begin{tabular}[c]{@{}c@{}}0.286\\ (0.18)\end{tabular}}}                      & \multicolumn{1}{c|}{{\color[HTML]{000000} \begin{tabular}[c]{@{}c@{}}0.004\\ (0.001)\end{tabular}}}                     & \multicolumn{1}{c|}{{\color[HTML]{000000} \begin{tabular}[c]{@{}c@{}}0.039\\ (0.015)\end{tabular}}}                     & \multicolumn{1}{c|}{{\color[HTML]{000000} \begin{tabular}[c]{@{}c@{}}0.023\\ (0.001)\end{tabular}}}                     & \multicolumn{1}{c|}{{\color[HTML]{000000} \begin{tabular}[c]{@{}c@{}}0.002\\ (0.003)\end{tabular}}}                                        & \multicolumn{1}{c|}{{\color[HTML]{000000} \begin{tabular}[c]{@{}c@{}}0.004\\ (0.003)\end{tabular}}}                                        & {\color[HTML]{000000} \begin{tabular}[c]{@{}c@{}}0.002\\ (0.001)\end{tabular}}                                        & {\color[HTML]{000000} } & \multicolumn{1}{c|}{{\color[HTML]{000000} \begin{tabular}[c]{@{}c@{}}0.55\\ (0.24)\end{tabular}}}                       & \multicolumn{1}{c|}{{\color[HTML]{000000} \begin{tabular}[c]{@{}c@{}}4.60\\ (1.96)\end{tabular}}}                       & \multicolumn{1}{c|}{{\color[HTML]{000000} \begin{tabular}[c]{@{}c@{}}23.6\\ (20.44)\end{tabular}}}                      & \multicolumn{1}{c|}{{\color[HTML]{000000} \begin{tabular}[c]{@{}c@{}}0.92\\ (0.06)\end{tabular}}}                       & \multicolumn{1}{c|}{{\color[HTML]{000000} \begin{tabular}[c]{@{}c@{}}1.20\\ (1.38)\end{tabular}}}                                          & \multicolumn{1}{c|}{{\color[HTML]{000000} \begin{tabular}[c]{@{}c@{}}7.25\\ (5.00)\end{tabular}}}                                          & {\color[HTML]{000000} \begin{tabular}[c]{@{}c@{}}1.73\\ (1.24)\end{tabular}}                                          &  & {\color[HTML]{000000} 9/45}                                                                                \\ \cline{1-8} \cline{10-16} \cline{18-18} 
{\color[HTML]{000000} \textbf{\begin{tabular}[c]{@{}c@{}}Sim\\ + SysID\end{tabular}}}                                & \multicolumn{1}{c|}{{\color[HTML]{000000} \begin{tabular}[c]{@{}c@{}}0.216\\ (0.13)\end{tabular}}}                      & \multicolumn{1}{c|}{{\color[HTML]{000000} \begin{tabular}[c]{@{}c@{}}0.002\\ (0.001)\end{tabular}}}                     & \multicolumn{1}{c|}{{\color[HTML]{000000} \begin{tabular}[c]{@{}c@{}}0.011\\ (0.008)\end{tabular}}}                     & \multicolumn{1}{c|}{{\color[HTML]{000000} \begin{tabular}[c]{@{}c@{}}0.001\\ (0.0001)\end{tabular}}}                    & \multicolumn{1}{c|}{{\color[HTML]{000000} \begin{tabular}[c]{@{}c@{}}0.001\\ (0.001)\end{tabular}}}                                        & \multicolumn{1}{c|}{{\color[HTML]{000000} \begin{tabular}[c]{@{}c@{}}0.001\\ (0.001)\end{tabular}}}                                        & {\color[HTML]{000000} \begin{tabular}[c]{@{}c@{}}0.001\\ (0.001)\end{tabular}}                                        & {\color[HTML]{000000} } & \multicolumn{1}{c|}{{\color[HTML]{000000} \begin{tabular}[c]{@{}c@{}}0.41\\ (0.21)\end{tabular}}}                       & \multicolumn{1}{c|}{{\color[HTML]{000000} \begin{tabular}[c]{@{}c@{}}1.63\\ (1.02)\end{tabular}}}                       & \multicolumn{1}{c|}{{\color[HTML]{000000} \begin{tabular}[c]{@{}c@{}}4.68\\ (4.90)\end{tabular}}}                       & \multicolumn{1}{c|}{{\color[HTML]{000000} \begin{tabular}[c]{@{}c@{}}0.05\\ (0.005)\end{tabular}}}                      & \multicolumn{1}{c|}{{\color[HTML]{000000} \begin{tabular}[c]{@{}c@{}}0.66\\ (0.6)\end{tabular}}}                                           & \multicolumn{1}{c|}{{\color[HTML]{000000} \begin{tabular}[c]{@{}c@{}}2.2\\ (1.96)\end{tabular}}}                                           & {\color[HTML]{000000} \begin{tabular}[c]{@{}c@{}}0.54\\ (0.5)\end{tabular}}                                           &  & {\color[HTML]{000000} 8/45}                                                                                \\ \cline{1-8} \cline{10-16} \cline{18-18} 
{\color[HTML]{000000} \textbf{\begin{tabular}[c]{@{}c@{}}Sim+SysID\\ +GP\end{tabular}}}                              & \multicolumn{1}{c|}{{\color[HTML]{000000} \begin{tabular}[c]{@{}c@{}}0.293\\ (0.35)\end{tabular}}}                      & \multicolumn{1}{c|}{{\color[HTML]{000000} \begin{tabular}[c]{@{}c@{}}0.001\\ (0.0004)\end{tabular}}}                    & \multicolumn{1}{c|}{{\color[HTML]{000000} \begin{tabular}[c]{@{}c@{}}0.010\\ (0.008)\end{tabular}}}                     & \multicolumn{1}{c|}{{\color[HTML]{000000} \begin{tabular}[c]{@{}c@{}}0.0009\\ (0.001)\end{tabular}}}                    & \multicolumn{1}{c|}{{\color[HTML]{000000} \begin{tabular}[c]{@{}c@{}}0.001\\ (0.002)\end{tabular}}}                                        & \multicolumn{1}{c|}{{\color[HTML]{000000} \begin{tabular}[c]{@{}c@{}}0.0005\\ (0.001)\end{tabular}}}                                       & {\color[HTML]{000000} \begin{tabular}[c]{@{}c@{}}0.001\\ (0.002)\end{tabular}}                                        & {\color[HTML]{000000} } & \multicolumn{1}{c|}{{\color[HTML]{000000} \begin{tabular}[c]{@{}c@{}}0.63\\ (0.53)\end{tabular}}}                       & \multicolumn{1}{c|}{{\color[HTML]{000000} \begin{tabular}[c]{@{}c@{}}1.09\\ (0.44)\end{tabular}}}                       & \multicolumn{1}{c|}{{\color[HTML]{000000} \begin{tabular}[c]{@{}c@{}}3.66\\ (3.51)\end{tabular}}}                       & \multicolumn{1}{c|}{{\color[HTML]{000000} \begin{tabular}[c]{@{}c@{}}0.039\\ (0.04)\end{tabular}}}                      & \multicolumn{1}{c|}{{\color[HTML]{000000} \begin{tabular}[c]{@{}c@{}}0.86\\ (0.9)\end{tabular}}}                                           & \multicolumn{1}{c|}{{\color[HTML]{000000} \begin{tabular}[c]{@{}c@{}}0.78\\ (1.16)\end{tabular}}}                                          & {\color[HTML]{000000} \begin{tabular}[c]{@{}c@{}}0.94\\ (0.92)\end{tabular}}                                          &  & {\color[HTML]{000000} 1/45}                                                                                \\ \cline{1-8} \cline{10-16} \cline{18-18} 
\end{tabular}
\vspace{-0.5em}
\end{table*}

\begin{table*}
\vspace{1em}
\centering
\caption{\textbf{Benchmark Results of Parameter Estimation In the Real-World.} The numbers represent the same meaning in Table \ref{table1}. In this experiment, we trained our method with a new dataset, where each inertial parameter was randomly sampled with a uniform distribution, independent of the object's shape. The OSI is trained with our SysID+GPs dataset. We employed TuneNet (Obs) with three iterations.} 
\label{table2}
\setlength{\tabcolsep}{1.3pt}
\begin{tabular}{
>{\columncolor[HTML]{EFEFEF}}c |ccccccc|
>{\columncolor[HTML]{FFFFFF}}c |ccccccc|c|cc}
\cline{1-8} \cline{10-16} \cline{18-19}
{\color[HTML]{000000} }                                                                                              & \multicolumn{7}{c|}{\cellcolor[HTML]{EFEFEF}{\color[HTML]{000000} Mean Absolute Error (MAE)}}                                                                                                                                                                                                                                                                                                                                                                                                                                                                                                                                                                                                                                                                                                                                                                                                                           & {\color[HTML]{000000} }                         & \multicolumn{7}{c|}{\cellcolor[HTML]{EFEFEF}{\color[HTML]{000000} Normalized Mean Absolute Error (NMAE)}}                                                                                                                                                                                                                                                                                                                                                                                                                                                                                                                                                                                                                                                                                                                                                                                                               &  & \cellcolor[HTML]{EFEFEF}{\color[HTML]{000000} }                                                                                 & \cellcolor[HTML]{EFEFEF}                                                             \\ \cline{1-8} \cline{10-16} \cline{18-19} 
\cellcolor[HTML]{EFEFEF}{\color[HTML]{000000} \textbf{\begin{tabular}[c]{@{}c@{}}Inertia\\ Parameters\end{tabular}}} & \multicolumn{1}{c|}{\cellcolor[HTML]{EFEFEF}{\color[HTML]{000000} \begin{tabular}[c]{@{}c@{}}Mass\\ (kg)\end{tabular}}} & \multicolumn{1}{c|}{\cellcolor[HTML]{EFEFEF}{\color[HTML]{000000} \begin{tabular}[c]{@{}c@{}}CoM x\\ (m)\end{tabular}}} & \multicolumn{1}{c|}{\cellcolor[HTML]{EFEFEF}{\color[HTML]{000000} \begin{tabular}[c]{@{}c@{}}CoM y\\ (m)\end{tabular}}} & \multicolumn{1}{c|}{\cellcolor[HTML]{EFEFEF}{\color[HTML]{000000} \begin{tabular}[c]{@{}c@{}}CoM z\\ (m)\end{tabular}}} & \multicolumn{1}{c|}{\cellcolor[HTML]{EFEFEF}{\color[HTML]{000000} \begin{tabular}[c]{@{}c@{}}Ixx\\ (kgm\textasciicircum{}2)\end{tabular}}} & \multicolumn{1}{c|}{\cellcolor[HTML]{EFEFEF}{\color[HTML]{000000} \begin{tabular}[c]{@{}c@{}}Iyy\\ (kgm\textasciicircum{}2)\end{tabular}}} & \cellcolor[HTML]{EFEFEF}{\color[HTML]{000000} \begin{tabular}[c]{@{}c@{}}Izz\\ (kgm\textasciicircum{}2)\end{tabular}} & {\color[HTML]{000000} }                         & \multicolumn{1}{c|}{\cellcolor[HTML]{EFEFEF}{\color[HTML]{000000} \begin{tabular}[c]{@{}c@{}}Mass\\ (kg)\end{tabular}}} & \multicolumn{1}{c|}{\cellcolor[HTML]{EFEFEF}{\color[HTML]{000000} \begin{tabular}[c]{@{}c@{}}CoM x\\ (m)\end{tabular}}} & \multicolumn{1}{c|}{\cellcolor[HTML]{EFEFEF}{\color[HTML]{000000} \begin{tabular}[c]{@{}c@{}}CoM y\\ (m)\end{tabular}}} & \multicolumn{1}{c|}{\cellcolor[HTML]{EFEFEF}{\color[HTML]{000000} \begin{tabular}[c]{@{}c@{}}CoM z\\ (m)\end{tabular}}} & \multicolumn{1}{c|}{\cellcolor[HTML]{EFEFEF}{\color[HTML]{000000} \begin{tabular}[c]{@{}c@{}}Ixx\\ (kgm\textasciicircum{}2)\end{tabular}}} & \multicolumn{1}{c|}{\cellcolor[HTML]{EFEFEF}{\color[HTML]{000000} \begin{tabular}[c]{@{}c@{}}Iyy\\ (kgm\textasciicircum{}2)\end{tabular}}} & \cellcolor[HTML]{EFEFEF}{\color[HTML]{000000} \begin{tabular}[c]{@{}c@{}}Izz\\ (kgm\textasciicircum{}2)\end{tabular}} &  & \multicolumn{1}{c|}{\cellcolor[HTML]{EFEFEF}{\color[HTML]{000000} \begin{tabular}[c]{@{}c@{}}Physical\\ Feasible\end{tabular}}} & \cellcolor[HTML]{EFEFEF}\begin{tabular}[c]{@{}c@{}}Inference\\ Time (s)\end{tabular} \\ \cline{1-8} \cline{10-16} \cline{18-19} 
{\color[HTML]{000000} \textbf{Ours}}                                                                                 & \multicolumn{1}{c|}{{\color[HTML]{000000} \begin{tabular}[c]{@{}c@{}}0.176\\ (0.16)\end{tabular}}}                      & \multicolumn{1}{c|}{{\color[HTML]{000000} \begin{tabular}[c]{@{}c@{}}0.013\\ (0.003)\end{tabular}}}                     & \multicolumn{1}{c|}{{\color[HTML]{000000} \begin{tabular}[c]{@{}c@{}}0.010\\ (0.009)\end{tabular}}}                     & \multicolumn{1}{c|}{{\color[HTML]{000000} \begin{tabular}[c]{@{}c@{}}0.026\\ (0.0007)\end{tabular}}}                    & \multicolumn{1}{c|}{{\color[HTML]{000000} \begin{tabular}[c]{@{}c@{}}0.002\\ (0.0009)\end{tabular}}}                                       & \multicolumn{1}{c|}{{\color[HTML]{000000} \begin{tabular}[c]{@{}c@{}}0.005\\ (0.0015)\end{tabular}}}                                       & {\color[HTML]{000000} \begin{tabular}[c]{@{}c@{}}0.002\\ (0.001)\end{tabular}}                                        & \cellcolor[HTML]{FFFFFF}{\color[HTML]{000000} } & \multicolumn{1}{c|}{{\color[HTML]{000000} \begin{tabular}[c]{@{}c@{}}0.357\\ (0.274)\end{tabular}}}                     & \multicolumn{1}{c|}{{\color[HTML]{000000} \begin{tabular}[c]{@{}c@{}}13.41\\ (3.803)\end{tabular}}}                     & \multicolumn{1}{c|}{{\color[HTML]{000000} \begin{tabular}[c]{@{}c@{}}2.749\\ (4.904)\end{tabular}}}                     & \multicolumn{1}{c|}{{\color[HTML]{000000} \begin{tabular}[c]{@{}c@{}}1.045\\ (0.03)\end{tabular}}}                      & \multicolumn{1}{c|}{{\color[HTML]{000000} \begin{tabular}[c]{@{}c@{}}1.352\\ (0.999)\end{tabular}}}                                        & \multicolumn{1}{c|}{{\color[HTML]{000000} \begin{tabular}[c]{@{}c@{}}10.37\\ (4.997)\end{tabular}}}                                        & {\color[HTML]{000000} \begin{tabular}[c]{@{}c@{}}1.307\\ (1.036)\end{tabular}}                                        &  & \multicolumn{1}{c|}{{\color[HTML]{000000} 3/45}}                                                                                & 0.102                                                                                \\ \cline{1-8} \cline{10-16} \cline{18-19} 
{\color[HTML]{000000} \textbf{OSI}}                                                                                  & \multicolumn{1}{c|}{{\color[HTML]{000000} \begin{tabular}[c]{@{}c@{}}0.243\\ (0.172)\end{tabular}}}                     & \multicolumn{1}{c|}{{\color[HTML]{000000} \begin{tabular}[c]{@{}c@{}}0.049\\ (0.011)\end{tabular}}}                     & \multicolumn{1}{c|}{{\color[HTML]{000000} \begin{tabular}[c]{@{}c@{}}0.032\\ (0.013)\end{tabular}}}                     & \multicolumn{1}{c|}{{\color[HTML]{000000} \begin{tabular}[c]{@{}c@{}}0.053\\ (0.019)\end{tabular}}}                     & \multicolumn{1}{c|}{{\color[HTML]{000000} \begin{tabular}[c]{@{}c@{}}0.052\\ (0.007)\end{tabular}}}                                        & \multicolumn{1}{c|}{{\color[HTML]{000000} \begin{tabular}[c]{@{}c@{}}0.041\\ (0.005)\end{tabular}}}                                        & {\color[HTML]{000000} \begin{tabular}[c]{@{}c@{}}0.075\\ (0.0067)\end{tabular}}                                       & {\color[HTML]{000000} }                         & \multicolumn{1}{c|}{{\color[HTML]{000000} \begin{tabular}[c]{@{}c@{}}0.478\\ (0.277)\end{tabular}}}                     & \multicolumn{1}{c|}{{\color[HTML]{000000} \begin{tabular}[c]{@{}c@{}}49.03\\ (11.89)\end{tabular}}}                     & \multicolumn{1}{c|}{{\color[HTML]{000000} \begin{tabular}[c]{@{}c@{}}18.55\\ (15.6)\end{tabular}}}                      & \multicolumn{1}{c|}{{\color[HTML]{000000} \begin{tabular}[c]{@{}c@{}}2.140\\ (0.785)\end{tabular}}}                     & \multicolumn{1}{c|}{{\color[HTML]{000000} \begin{tabular}[c]{@{}c@{}}29.41\\ (12.27)\end{tabular}}}                                        & \multicolumn{1}{c|}{{\color[HTML]{000000} \begin{tabular}[c]{@{}c@{}}80.48\\ (32.68)\end{tabular}}}                                        & {\color[HTML]{000000} \begin{tabular}[c]{@{}c@{}}44.76\\ (19.81)\end{tabular}}                                        &  & \multicolumn{1}{c|}{{\color[HTML]{000000} 0/45}}                                                                                & 0.100                                                                                \\ \cline{1-8} \cline{10-16} \cline{18-19} 
{\color[HTML]{000000} \textbf{\begin{tabular}[c]{@{}c@{}}TuneNet\\ (Obs) \cite{allevato2020tunenet}\end{tabular}}}                              & \multicolumn{1}{c|}{{\color[HTML]{000000} \begin{tabular}[c]{@{}c@{}}0.3\\ (0.2)\end{tabular}}}                         & \multicolumn{1}{c|}{{\color[HTML]{000000} \begin{tabular}[c]{@{}c@{}}0.004\\ (0.0001)\end{tabular}}}                    & \multicolumn{1}{c|}{{\color[HTML]{000000} \begin{tabular}[c]{@{}c@{}}0.015\\ (0.007)\end{tabular}}}                     & \multicolumn{1}{c|}{{\color[HTML]{000000} \begin{tabular}[c]{@{}c@{}}0.025\\ (0.0004)\end{tabular}}}                    & \multicolumn{1}{c|}{{\color[HTML]{000000} \begin{tabular}[c]{@{}c@{}}0.0008\\ (0.0004)\end{tabular}}}                                      & \multicolumn{1}{c|}{{\color[HTML]{000000} \begin{tabular}[c]{@{}c@{}}0.0007\\ (0.0006)\end{tabular}}}                                      & {\color[HTML]{000000} \begin{tabular}[c]{@{}c@{}}0.0008\\ (0.0001)\end{tabular}}                                      & {\color[HTML]{000000} }                         & \multicolumn{1}{c|}{{\color[HTML]{000000} \begin{tabular}[c]{@{}c@{}}0.744\\ (0.490)\end{tabular}}}                     & \multicolumn{1}{c|}{{\color[HTML]{000000} \begin{tabular}[c]{@{}c@{}}4.828\\ (0.15)\end{tabular}}}                      & \multicolumn{1}{c|}{{\color[HTML]{000000} \begin{tabular}[c]{@{}c@{}}9.425\\ (7.704)\end{tabular}}}                     & \multicolumn{1}{c|}{{\color[HTML]{000000} \begin{tabular}[c]{@{}c@{}}1.020\\ (0.016)\end{tabular}}}                     & \multicolumn{1}{c|}{{\color[HTML]{000000} \begin{tabular}[c]{@{}c@{}}0.368\\ (0.196)\end{tabular}}}                                        & \multicolumn{1}{c|}{{\color[HTML]{000000} \begin{tabular}[c]{@{}c@{}}1.474\\ (0.859)\end{tabular}}}                                        & {\color[HTML]{000000} \begin{tabular}[c]{@{}c@{}}0.355\\ (0.16)\end{tabular}}                                         &  & \multicolumn{1}{c|}{{\color[HTML]{000000} 0/45}}                                                                                & 7-8                                                                                  \\ \cline{1-8} \cline{10-16} \cline{18-19} 
{\color[HTML]{000000} \textbf{DROPO \cite{tiboni2023dropo}}}                                                                                & \multicolumn{1}{c|}{{\color[HTML]{000000} \begin{tabular}[c]{@{}c@{}}0.34\\ (0.21)\end{tabular}}}                       & \multicolumn{1}{c|}{{\color[HTML]{000000} \begin{tabular}[c]{@{}c@{}}0.009\\ (0.001)\end{tabular}}}                     & \multicolumn{1}{c|}{{\color[HTML]{000000} \begin{tabular}[c]{@{}c@{}}0.014\\ (0.009)\end{tabular}}}                     & \multicolumn{1}{c|}{{\color[HTML]{000000} \begin{tabular}[c]{@{}c@{}}0.026\\ (0.009)\end{tabular}}}                     & \multicolumn{1}{c|}{{\color[HTML]{000000} \begin{tabular}[c]{@{}c@{}}0.0025\\ (0.0008)\end{tabular}}}                                      & \multicolumn{1}{c|}{{\color[HTML]{000000} \begin{tabular}[c]{@{}c@{}}0.0025\\ (0.0001)\end{tabular}}}                                      & {\color[HTML]{000000} \begin{tabular}[c]{@{}c@{}}0.0024\\ (0.0009)\end{tabular}}                                      & {\color[HTML]{000000} }                         & \multicolumn{1}{c|}{{\color[HTML]{000000} \begin{tabular}[c]{@{}c@{}}0.965\\ (0.889)\end{tabular}}}                     & \multicolumn{1}{c|}{{\color[HTML]{000000} \begin{tabular}[c]{@{}c@{}}9.812\\ (1.031)\end{tabular}}}                     & \multicolumn{1}{c|}{{\color[HTML]{000000} \begin{tabular}[c]{@{}c@{}}6.056\\ (4.428)\end{tabular}}}                     & \multicolumn{1}{c|}{{\color[HTML]{000000} \begin{tabular}[c]{@{}c@{}}1.061\\ (0.380)\end{tabular}}}                     & \multicolumn{1}{c|}{{\color[HTML]{000000} \begin{tabular}[c]{@{}c@{}}1.444\\ (0.969)\end{tabular}}}                                        & \multicolumn{1}{c|}{{\color[HTML]{000000} \begin{tabular}[c]{@{}c@{}}4.697\\ (4.39)\end{tabular}}}                                         & {\color[HTML]{000000} \begin{tabular}[c]{@{}c@{}}1.438\\ (1.014)\end{tabular}}                                        &  & \multicolumn{1}{c|}{{\color[HTML]{000000} 11/45}}                                                                               & 10 \textless{}                                                                       \\ \cline{1-8} \cline{10-16} \cline{18-19} 
\end{tabular}
\vspace{-1em}
\end{table*}

\begin{table}[t]
\centering
\caption{\textbf{Runtime Benchmark in Simulation.} 
All learning-based methods demonstrated rapid estimation speeds, with runtimes around 0.1 second. Traditional methods take much time due to the persistently excitation signal.}
\label{table3}
\setlength{\tabcolsep}{3pt}
\begin{tabular}{c|c|c|c|c|c|c|c}
\hline
\rowcolor[HTML]{EFEFEF} 
\cellcolor[HTML]{EFEFEF}{\color[HTML]{000000} \textbf{Method}}   & {\color[HTML]{000000} OLS(S)} & {\color[HTML]{000000} WLS(S)} & {\color[HTML]{000000} OLS(L)} & {\color[HTML]{000000} WLS(L)} & {\color[HTML]{000000} LSTM}  & {\color[HTML]{000000} TCN}   & {\color[HTML]{000000} 1D-CNN} \\ \hline
\cellcolor[HTML]{EFEFEF}{\color[HTML]{000000} \textbf{Time (s)}} & {\color[HTML]{000000} 0.525}  & {\color[HTML]{000000} 0.56}   & {\color[HTML]{000000} 10.09}  & {\color[HTML]{000000} 10.4}   & {\color[HTML]{000000} 0.102} & {\color[HTML]{000000} 0.105} & {\color[HTML]{000000} 0.102}  \\ \hline
\end{tabular}
\vspace{-2em}
\end{table}

\subsubsection{Estimation Results with Real World Dataset}
In the real-world estimation of inertial parameters (see Table \ref{table1}), both \textbf{Sim+SysID} and \textbf{Sim+SysID+GP} models exhibit superior estimation performance, underscoring the effectiveness of real-to-sim adaptation. While the results are not as high as those achieved in simulations (shown in Fig. \ref{estimation_benchmark_sim}), they surpass traditional methods in estimating the CoM and inertia. In contrast, the \textbf{PureSim}, \textbf{Sim+GP}, and \textbf{Sim+ActNet} models demonstrate overfitting to simulation data, which is influenced by their limited sim-to-real adaptation ability. (see Fig. \ref{sim2real_fig2}). We observed that using a GP slightly improves estimation performance compared to \textbf{Sim+SysID}, likely because GPs are not trained with specific object information, limiting their generalization. Additionally, the noise inherent in the signals captured by the GPs could impede the extraction of crucial features needed for accurate estimation. Despite applying soft regularization to ensure physically feasible solutions, 44 out of 45 solutions of \textbf{Sim+SysID+GP} meet the criteria for physical consistency, whereas most solutions from existing methods using non-constrained optimization did not satisfy physical consistency. 

\subsubsection{Benchmark Results Using More Generalizable Dataset}
We reported benchmark performance comparison results in Table \ref{table3}. Our proposed has no noticeable decrease in performance even when trained on a more generalizable dataset with uniformly and independently selected parameters. Training with a uniform distribution for each parameter independently helps estimate mass due to its wide coverage, but is less effective for estimating the center of mass (CoM) and inertia, which are more influenced by the object's shape. Our method achieves comparable accuracy to baselines like TuneNet and DROPO, which require multiple iterations for parameter estimation, but ours operates significantly faster. While additional iterations can enhance accuracy, they also increase inference time, rendering them unsuitable for real-time applications. Since we handle the adaptation entirely offline and separate it from the actual evaluation step, it provides advantages in both speed and accuracy. Specifically, we reduce the \textit{reality gap} of the robot system offline and then train a data-driven estimator on this adapted model. In contrast, other methods iteratively tune simulation parameters to reduce the \textit{reality gap}.

\subsection{Control Task Experiments}
\subsubsection{Manipulation Task}
After estimating an object's inertial parameters, we use these parameters for gravity compensation, effectively demonstrating the benefits of explicit parameter utilization. As illustrated in Fig. \ref{experiments_task_fig}, there is a notable 36\% performance enhancement along the y-axis, where gravity is exerted. Transitioning to a controller based on impedance control or inverse dynamics, which rely more heavily on model accuracy, promises even greater performance improvements in handling dynamic environments and interactions. 

\subsubsection{Locomotion Task}
We integrated the estimated parameters into the baseline controller by recalculating the SATYRR robot's equilibrium point, considering the combined CoM and mass. This updated equilibrium point was used as the desired pitch angle in an LQR controller, resulting in a 65\% improvement in position tracking (Fig. \ref{experiments_task_fig}). Even greater performance gains can be expected with model-predictive or whole-body controllers, which rely more on model accuracy.


\begin{figure}[t]
    \centering
    \begin{subfigure}[b]{0.3\linewidth}
        \includegraphics[width=\columnwidth]{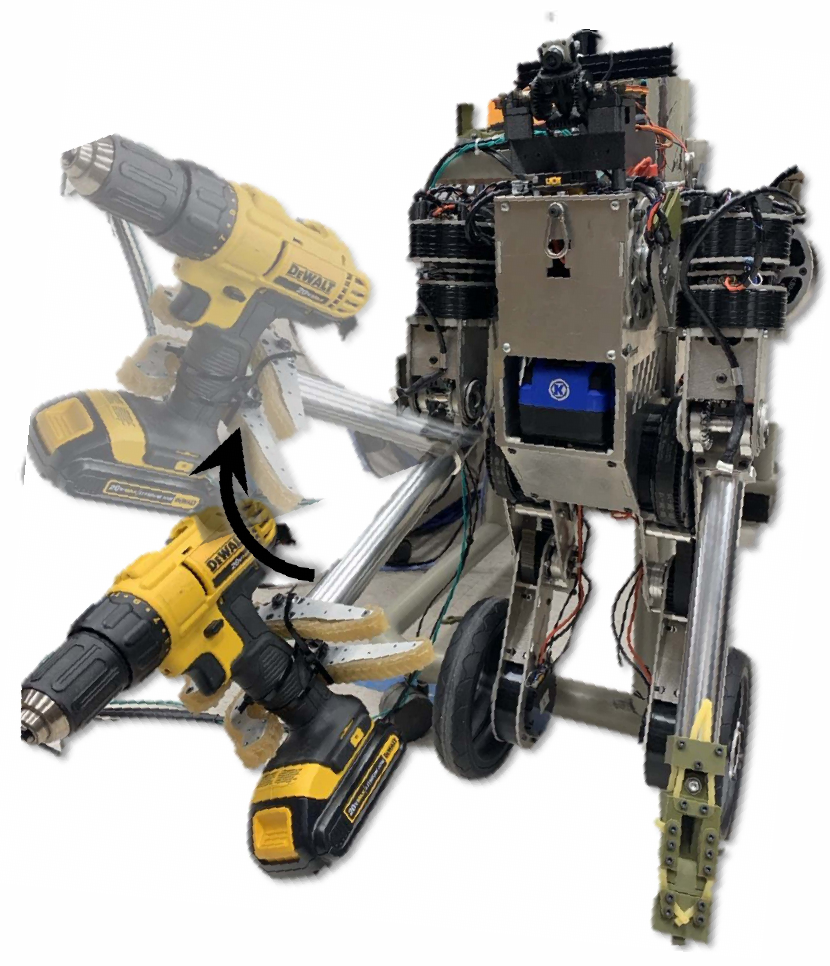}
            \caption{Manipulation task with holding a drill}
    \end{subfigure} 
        \hspace{0.05cm}
    \begin{subfigure}[b]{0.45\linewidth}
        \includegraphics[width=\columnwidth]{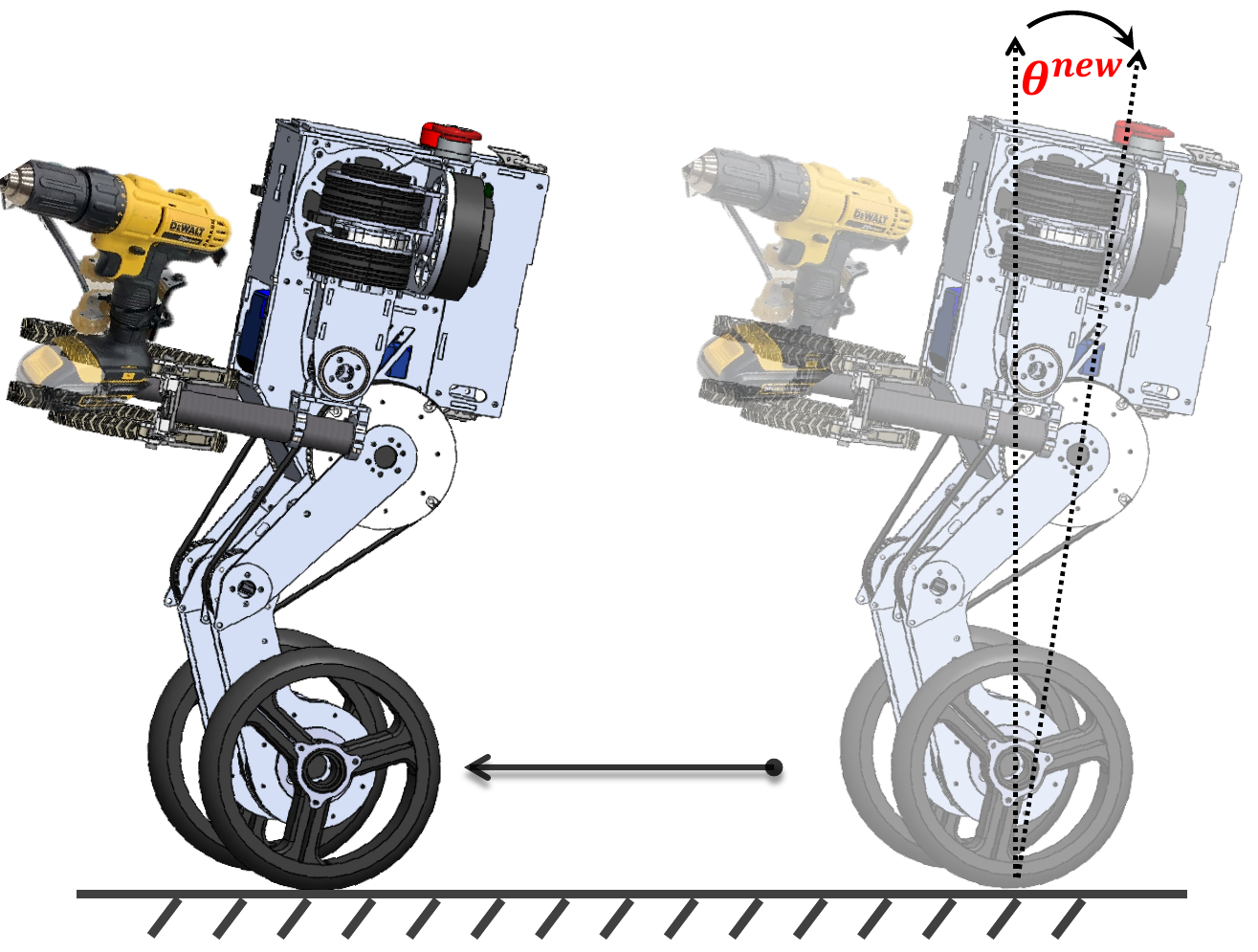}
        \caption{Locomotion task with holding a drill}
    \end{subfigure}
\caption{\textbf{Control Task Experiments}. The recorded motion obtained from a Human machine interface (HMI) is used as a desired trajectory in a manipulation task. The forward and backward driving tests were conducted using HMI (refer to the attached video).}
\label{task_fig}
\vspace{-1em}
\end{figure}

\begin{figure}[t]
     \centering
 	\begin{subfigure}[b]{0.425\linewidth}
 		\includegraphics[width=\columnwidth]{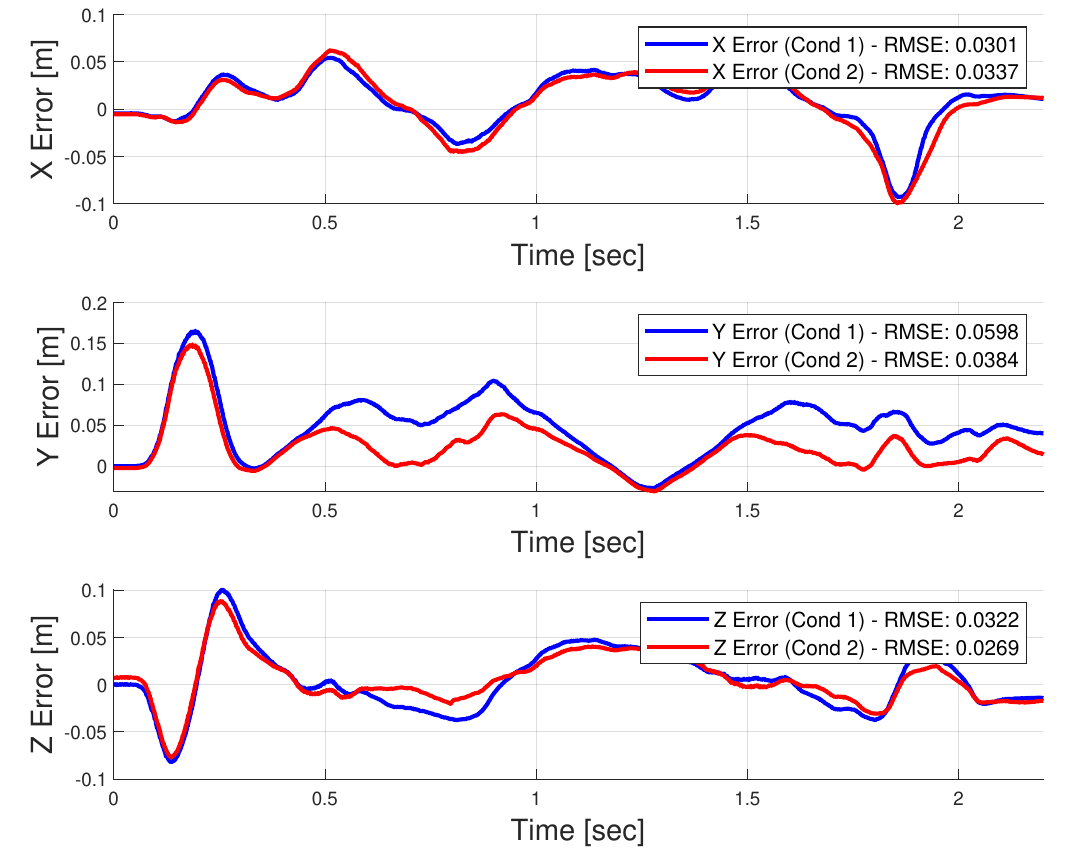}
            \caption{Manipulation Task Result}
 	\end{subfigure} 
 	\begin{subfigure}[b]{0.56\linewidth}
 		\includegraphics[width=\columnwidth]{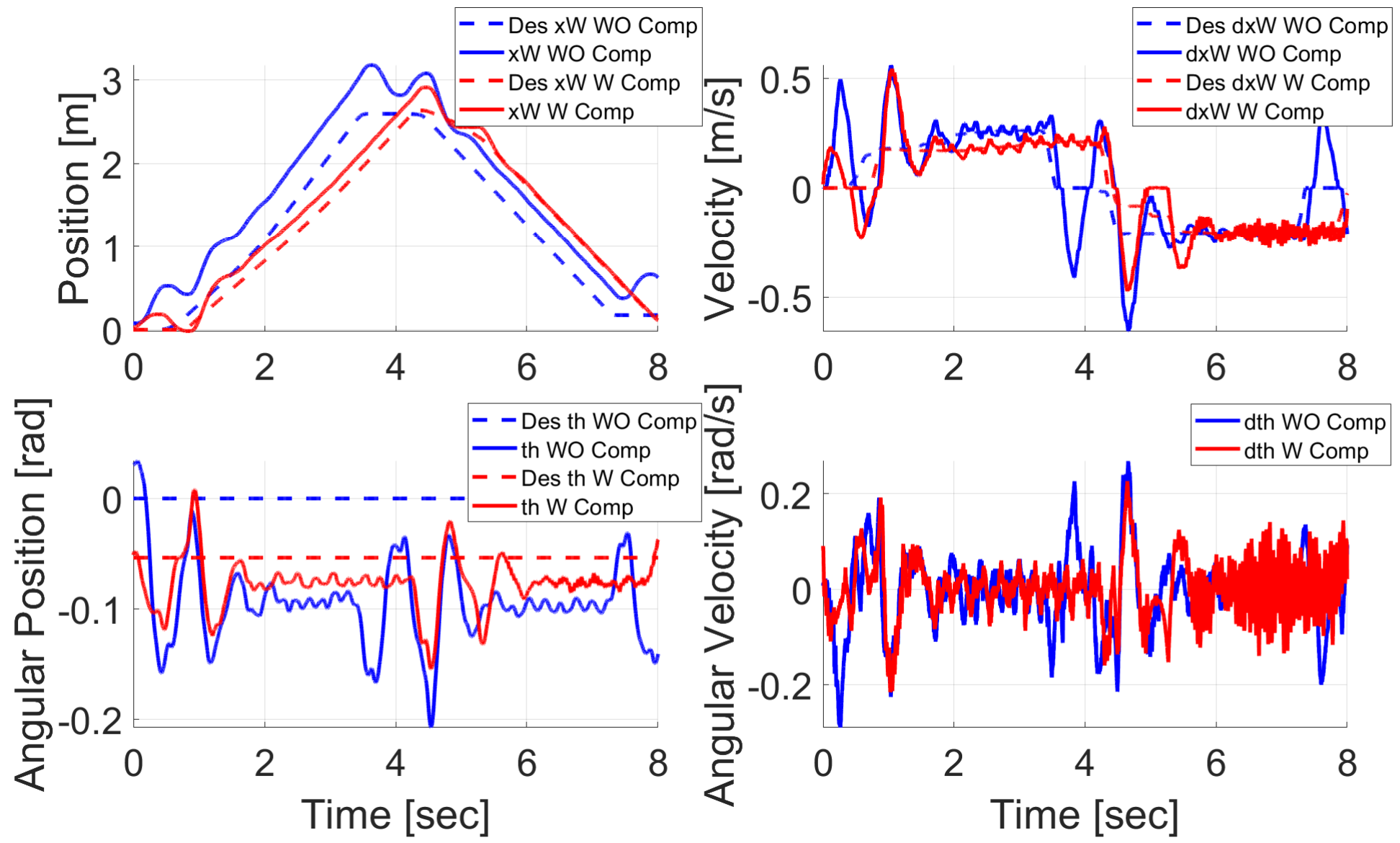}
 		\caption{Locomotion Task Result}
 	\end{subfigure}
\caption{\textbf{Results of Task Experiments}. Integrating the estimated inertial parameters of an object with a baseline controller significantly improves tracking performance for both manipulation and locomotion tasks. The estimated parameters are as follows: $\br{y_S}=[1.1, -0.008, 0.1, 0.025, 0.005, 0.0007, 0.005]^\tp$.}
\label{experiments_task_fig}
\vspace{-1.5em}
\end{figure}

%% file: 7_conclusion.tex
\section{CONCLUSION}
\label{c:1}

In this paper, we propose a fast online learning-based framework to identify the inertial parameters of unknown objects, enhancing the accuracy of model-based controllers and making them more suitable for real-time applications. We introduce a real-to-sim adaptation that combines \textit{Robot System Identification} and \textit{Gaussian Processes} to reduce the \textit{reality gap} caused by parametric and non-parametric modeling error, respectively. The adaptation method maintained effective performance even with new objects attached, and our estimation framework is significantly faster than other methodologies. Exploring the impact of accurate system parameter estimation on RL policy would be an interesting future research direction.

